\documentclass[letterpaper]{article}
\usepackage[margin=1in,dvips]{geometry}
\usepackage{graphicx,psfrag,amsmath,amsthm,amssymb}
\usepackage{natbib}
\usepackage{algorithmic,algorithm}
\usepackage{times}
\usepackage{url}
\usepackage{hyperref}

\newcommand{\A}{\ensuremath{\mathbf{A}}}

\newcommand{\D}{\ensuremath{\mathbf{D}}}

\newcommand{\I}{\ensuremath{\mathbf{I}}}

\newcommand{\K}{\ensuremath{\mathbf{K}}}

\newcommand{\Q}{\ensuremath{\mathbf{Q}}}
\newcommand{\RR}{\ensuremath{\mathbf{R}}}
\renewcommand{\SS}{\ensuremath{\mathbf{S}}}
\newcommand{\T}{\ensuremath{\mathbf{T}}}
\newcommand{\U}{\ensuremath{\mathbf{U}}}
\newcommand{\V}{\ensuremath{\mathbf{V}}}
\newcommand{\W}{\ensuremath{\mathbf{W}}}

\newcommand{\f}{\ensuremath{\mathbf{f}}}
\newcommand{\g}{\ensuremath{\mathbf{g}}}

\newcommand{\w}{\ensuremath{\mathbf{w}}}
\newcommand{\x}{\ensuremath{\mathbf{x}}}
\newcommand{\y}{\ensuremath{\mathbf{y}}}

\newcommand{\bSigma}{\ensuremath{\boldsymbol{\Sigma}}}

\newcommand{\bbE}{\ensuremath{\mathbb{E}}}

\newcommand{\bbR}{\ensuremath{\mathbb{R}}}

\newcommand{\calA}{\ensuremath{\mathcal{A}}}
\newcommand{\calB}{\ensuremath{\mathcal{B}}}

\newcommand{\calH}{\ensuremath{\mathcal{H}}}

\newcommand{\calK}{\ensuremath{\mathcal{K}}}

\newcommand{\calS}{\ensuremath{\mathcal{S}}}

\newcommand{\norm}[1]{\left\lVert#1\right\rVert}

{%
\begin{list}{#1}{
\vspace{-\topsep}
\vspace{-\partopsep}
\setlength{\itemindent}{0cm}
\setlength{\rightmargin}{0cm}
\setlength{\listparindent}{0cm}
\settowidth{\labelwidth}{#1}
\setlength{\leftmargin}{\labelwidth}
\addtolength{\leftmargin}{\labelsep}
\setlength{\itemsep}{0cm}
}%
}%
{%
\end{list}
\vspace{-\topsep}
\vspace{-\partopsep}
}

{\begin{enumerate}%
}%
{\end{enumerate}}

\theoremstyle{plain}% default
\newtheorem{thm}{Theorem}[section]
\newtheorem{lemma}[thm]{Lemma}
\newtheorem*{lemma*}{Lemma}

\newtheorem*{prop*}{Proposition}

\theoremstyle{definition}

\newtheorem*{defn*}{Definition}

\newtheorem*{exmp*}{Example}

\newtheorem*{conj*}{Conjecture}

\theoremstyle{remark}

\newtheorem*{rmk*}{Remark}

\graphicspath{{grf/}}

\newcommand{\expx}[1]{\bbE\!\left[#1\right]}
\newcommand{\expy}[1]{\bbE\!\left[#1\right]}
\newcommand{\expxy}[1]{\bbE\!\left[#1\right]}
\newcommand{\expxi}[1]{\bbE[#1]} % inline
\newcommand{\expyi}[1]{\bbE[#1]} % inline
\newcommand{\expxyi}[1]{\bbE[#1]} % inline
\newcommand{\ie}{{\em i.e., }}
\newcommand{\eg}{{\em e.g., }}

\renewcommand{\calA}{\ensuremath{\boldsymbol{\mathcal{A}}}}
\renewcommand{\calB}{\ensuremath{\boldsymbol{\mathcal{B}}}}

\newcommand{\kl}[1]{}
\newcommand{\wwcomment}[1]{}
\newcommand{\tmcomment}[1]{}

\title{Nonparametric Canonical Correlation Analysis}

\author{Tomer Michaeli$^1$ \hspace{2em}  Weiran Wang$^2$ \hspace{2em} Karen Livescu$^2$ \\
  $^1$Technion--Israel Institute of Technology, Haifa, Israel \\
  $^2$TTI-Chicago, Chicago, IL 60637, USA \\
  \texttt{tomer.m@technion.ac.il} \hspace{1em} \texttt{\{weiranwang,klivescu\}@ttic.edu}
}

\begin{document}
\maketitle 

\begin{abstract}

\kl{did a little rewording}

Canonical correlation analysis (CCA) is a classical representation learning technique for finding correlated variables in multi-view data.
Several nonlinear extensions of the original linear CCA have been proposed, including kernel and deep neural network methods. %These approaches restrict attention to certain function families specified by the user (by choosing a kernel or neural network structure) and are computationally demanding.
These approaches seek maximally correlated projections among families of functions, which the user specifies (by choosing a kernel or neural network structure), and are computationally demanding. Interestingly, the theory of nonlinear CCA, without functional restrictions, had been studied in the population setting by Lancaster already in the 1950s, but these results have not inspired practical algorithms. We revisit Lancaster's theory to devise a practical algorithm for nonparametric CCA (NCCA). Specifically, we show that the solution can be expressed in terms of the singular value decomposition of a certain operator associated with the joint density of the views. Thus, by estimating the population density from data, NCCA reduces to solving an eigenvalue system, superficially like kernel CCA but, importantly, without requiring the inversion of any kernel matrix. We also derive a partially linear CCA (PLCCA) variant in which one of the views undergoes a linear projection while the other is nonparametric.
%PLCCA turns out to have a similar form to the linear CCA, but with a nonparametric regression term replacing the linear regression in CCA.
Using a kernel density estimate based on a small number of nearest neighbors, our NCCA and PLCCA algorithms are memory-efficient, often run much faster, and %achieve better performance than kernel CCA and comparable performance to deep CCA.
perform better than kernel CCA and comparable to deep CCA.

\end{abstract}

\section{Introduction}

A common task in data analysis is to reveal the common variability in multiple views of the same phenomenon, while suppressing view-specific noise factors.  Canonical correlation analysis (CCA)~\citep{Hotell36a} is a classical statistical technique that targets this goal. In CCA, linear projections of two random vectors are sought, such that the resulting low-dimensional vectors are maximally correlated. This tool has found widespread use in various fields, including
%meteorology \citet{Anderson58}, chemometrics \citet{Montanarella95},
recent application to natural language processing~\citep{Dhillon_11b}, speech recognition \citep{AroraLivesc13a}, genomics \citep{Witten09}, and cross-modal retrieval~\citep{gong2014improving}. %\klcomment{Changed this to only cite recent work, because it saves references and seems more relevant for the NIPS community.}
%to detect interesting phenomena shared by two data sets and to learn informative single-view features based on multi-view training data. Examples include

One of the shortcomings of CCA is its restriction to linear mappings, since many real-world multi-view datasets exhibit highly nonlinear relationships. To overcome this limitation, several extensions of CCA have been proposed for finding maximally correlated \emph{nonlinear} projections. In kernel CCA (KCCA) \citep{Akaho01a,Melzer_01a,BachJordan02a,Hardoon_04a}, these nonlinear mappings are chosen from two reproducing kernel Hilbert spaces (RKHS). In deep CCA (DCCA) \citep{Andrew_13a}, the projections are obtained from two deep neural networks that are trained to output maximally correlated vectors. Nonparametric CCA-type methods, which are not limited to specific function classes, include the alternating conditional expectations (ACE) algorithm and its extensions \citep{Breiman85,Balakr_12a,Makur_15a}. Nonlinear CCA methods are advantageous over linear CCA in a range of applications \citep{Hardoon_04a,
%Vinokour_03a,
%Dhillon_11b,Witten09,
Melzer_01a,Wang_15b}.  However, existing nonlinear CCA approaches are very computationally demanding, and are often impractical to apply on large data.

Interestingly, the problem of finding the most correlated nonlinear projections of two random variables has been studied by \citet{lancaster1958} and \citet{hannan1961}, long before the derivation of KCCA, DCCA and ACE. They characterized the optimal projections in the population setting, without restricting the solution to an RKHS or to have any particular parametric form. However, these theoretical results have not inspired practical algorithms.

In this paper, we revisit Lancaster's theory, and use it to devise a practical algorithm for \emph{nonparametric CCA} (NCCA). Specifically, we show that the solution to the nonlinear CCA problem can be expressed in terms of the singular value decomposition (SVD) of a certain operator, which is defined via the population density. Therefore, to obtain a practical method, we estimate the density from training data and use the estimate in the solution. The resulting algorithm reduces to solving an eigenvalue system with a particular kernel that depends on the joint distribution between the views.  While superficially similar to other eigenvalue methods, it is fundamentally different from them and in particular has crucial advantages over KCCA.  For example, unlike KCCA, NCCA does not involve computing the inverse of any matrix, making it computationally feasible on large data where KCCA (even using approximation techniques) is impractical.  We elucidate this and other contrasts in Sec.~\ref{sec:algorithm} below.  We show that NCCA achieves state-of-the art performance, while being much more computationally efficient than KCCA and DCCA.

In certain situations, nonlinearity is needed for one view but not for the other. In such cases, it may be advantageous to constrain the projection of the second view to be linear. An additional contribution of this paper is the derivation of a closed-form solution to this \emph{partially linear CCA} (PLCCA) problem in the population setting. We show that PLCCA has essentially the same form as linear CCA, but with the optimal linear predictor term in CCA replaced by an optimal nonlinear predictor in PLCCA. Thus, moving from the population setting to sample data entails simply using nonlinear regression to estimate this predictor. The resulting algorithm is efficient and, as we demonstrate on realistic data, sometimes matches DCCA and significantly outperforms CCA and KCCA.
\section{Background}
\label{sec:backgound}

We start by reviewing the original CCA algorithm \citep{Hotell36a}. Let $X\in\bbR^{D_x}$ and $Y\in\bbR^{D_y}$ be two random vectors (views). The goal in CCA is to find a pair of $L$-dimensional projections $\W_1^\top X$, $\W_2^\top Y$ that are maximally correlated, but where different dimensions within each view are constrained to be uncorrelated.  Assuming for notational simplicity that $X$ and $Y$ have zero mean, the CCA problem can be written as\footnote{Here and throughout, expectations are with respect to the joint distribution of all random variables (capital letters) appearing within the square brackets of the expectation operator $\bbE$.}
%CCA can be extended to more than one projection, by maximizing the sum of correlations in the individual directions, subject to the constraint that successive projections are uncorrelated with each other.  Stacking the $L$ projection directions of the two views as two matrices $\W_1\in\bbR^{D_x\times L}$ and $\W_2\in\bbR^{D_y\times L}$, the multi-dimensional CCA problem becomes %\wwcomment{removed dimensions of $\W_1$ and $\W_2$ from $\max$ to fit the equation}
\begin{gather} \label{e:ccaL}
  \max_{\W_1,\W_2} \expxy{\left(\W_1^\top X\right)^\top\!\left(\W_2^\top Y\right)} \\
%  \max_{\substack{\W_1\in\bbR^{D_x\times L}\\ \W_2\in\bbR^{D_y\times L}}} \expxy{\left(\W_1^\top X\right)^\top\!\left(\W_2^\top Y\right)}
  \text{s.t.} \; %\W_1^\top \bSigma_{xx} \W_1 = \W_2^\top \bSigma_{yy} \W_2 = \I.
  \expx{\left(\W_1^\top X\right)\!\left(\W_1^\top X\right)^\top}\!\!=\expy{\left(\W_2^\top Y\right)\!\left(\W_2^\top Y\right)^\top}\!\!=\I, \nonumber
\end{gather}
where the maximization is over $\W_1\in\bbR^{D_x\times L},\W_2\in\bbR^{D_y\times L}$.
%this is equivalent to finding the directions $(\w_1,\w_2)$ that maximize the covariance $\expxyi{(\w_1^\top X) (\w_2^\top Y)}$, subject to the restriction that the projections have unit variance, $\expxi{(\w_1^\top X)^2}=\expxi{(\w_1^\top Y)^2}=1$.
This objective has been extensively studied and is known to be optimal in several senses: It maximizes the mutual information for certain distributions $p(\x,\y)$ \citep{Borga01a}, maximizes the likelihood for certain latent variable models \citep{BachJordan05a}, and is equivalent to the information bottleneck method when $p(\x,\y)$ is Gaussian \citep{Chechik_05a}.

%\begin{align} \label{e:ncca}
%  \max_{\substack{\U\in\bbR^{D_y\times L}\\ \V\in\bbR^{D_x\times L}}} \expxy{\tilde{\x}^\top\tilde{\y}} \;\; \text{s.t.} \;\; \expx{\tilde{\x}\tilde{\x}^\top}=\expy{\tilde{\y}\tilde{\y}^\top}=\I.
%\end{align}
%where $\tilde{\x}=\U^\top\x$ and $\tilde{\y}=\V^\top\y$.
%\begin{gather} \label{e:cca}
%  \max\limits_{\uu\in\bbR^{D_x},\vv\in\bbR^{D_y}} \; \frac{\expxy{(\uu^\top \x) (\vv^\top \y)}}{ \sqrt{\expx{(\uu^\top \x)^2}}\sqrt{\expy{(\vv^\top \y)^2}} } =
%  \frac{\uu^\top \bSigma_{xy} \vv }{(\uu^\top \bSigma_{xx} \uu) (\vv^\top \bSigma_{yy} \vv) }.
%\end{gather}
%In practice, we do not have access to the joint distributions $\bbP(\x,\y)$ or the marginal distributions $\bbP(\x)$ or $\bbP(\y)$, so the covariances are estimated empirically using the training samples, i.e., $\bSigma_{xy}=\expxy{\x\y^\top}\approx \frac{1}{N} \sum_{i=1}^N \x_i \y_i^\top$ and similarly for $\bSigma_{xx}$ and $\bSigma_{yy}$.

%The optimal projections depend only on the second-order statistics of the views, captured by the matrices $\bSigma_{xx}=\expxi{XX^\top}$, $\bSigma_{yy}=\expyi{YY^\top}$ and $\bSigma_{xy}=\expxyi{XY^\top}$. Specifically, t
The CCA solution can be expressed as $(\W_1,\W_2)=(\bSigma_{xx}^{-1/2}\U,\bSigma_{yy}^{-1/2}\V)$, where $\bSigma_{xx}=\expxi{XX^\top}$, $\bSigma_{yy}=\expyi{YY^\top}$,  $\bSigma_{xy}=\expxyi{XY^\top}$, and $\U\in\bbR^{D_x\times L}$ and $\V\in\bbR^{D_y\times L}$ are the top $L$ left and right singular vectors of the matrix $\T=\bSigma_{xx}^{-1/2}\bSigma_{xy}\bSigma_{yy}^{-1/2}$ (see \citep{mardia_79}). In practice, the joint distribution $p(\x,\y)$ is rarely known, and only paired multi-view samples $\{(\x_i,\y_i)\}_{i=1}^N$ are available, so the population covariances are replaced by their empirical estimates.\footnote{$\bSigma_{xy}\approx \frac{1}{N} \sum_{i=1}^N \x_i \y_i^\top$ and similarly for $\bSigma_{xx}$ and $\bSigma_{yy}$.}

To facilitate the analogy with partially linear CCA (Sec.~\ref{sec:PLCCA}), we note that the CCA solution can also be expressed in terms of the optimal predictor % (TM)
%best linear predictor
% (KL)
%<<<<<<< .mine
%%(in the mean squared error sense)
%of $X$ from $Y$, which is given by $\hat{X}=\bSigma_{xy}\bSigma_{yy}^{-1}Y$, and
%in terms of its covariance $\bSigma_{\hat{x}\hat{x}}=\bSigma_{xy}\bSigma_{yy}^{-1}\bSigma_{yx}$.  Specifically, $\U$ corresponds to the eigenvectors of $\K=\T\T^\top=\bSigma_{xx}^{-1/2}\bSigma_{\hat{x}\hat{x}}\bSigma_{xx}^{-1/2}$,
%=======
(in the mean squared error sense) of $X$ from $Y$, given by $\hat{X}=\bSigma_{xy}\bSigma_{yy}^{-1}Y$, and
%in terms of
its covariance $\bSigma_{\hat{x}\hat{x}}=\bSigma_{xy}\bSigma_{yy}^{-1}\bSigma_{yx}$.  Specifically, $\U$ corresponds to the eigenvectors of $\K=\T\T^\top=\bSigma_{xx}^{-1/2}\bSigma_{\hat{x}\hat{x}}\bSigma_{xx}^{-1/2}$,
%>>>>>>> .r54804
% and thus it can be verified that
and, by algebraic manipulation, the optimal projections can be written as
\begin{equation}\label{e:ccaL2}
\W_1^\top X = \U^\top \bSigma_{xx}^{-\frac{1}{2}} X,\quad \W_2^\top Y = \D^{-\frac{1}{2}} \U^\top \bSigma_{xx}^{-\frac{1}{2}} \hat{X},
\end{equation}
where $\D$ is a diagonal matrix with the top $L$ eigenvalues of $\K$ on its diagonal.

%\subsection{Nonlinear CCA extensions}
Since the representation power of linear mappings is limited, several nonlinear extensions of problem~\eqref{e:ccaL} have been proposed. These methods find two maximally correlated \emph{nonlinear} projections %\kl{use boldface for multidimensional f, g?} \tmcomment{Not sure... if $f,g$ are bold, then what would you use for their components $f_i,g_i$? they are scalar functions. I think its fine the way it is now} \kl{I am fine with it but the convention I usually use is bold for all vectors and non-bold for all scalars.  So ${\mathbf f}$ is a vector with scalar components $f_i$ (and ${\mathbf f_i}$ would be, e.g., vector elements of a set)} \tmcomment{OK, I switched to bold}
%\wwcomment{I was fine with the previous notations (non-boldface). If we switch to boldface, there are some inconsistency now in the paper, see my comments. Shall we have a ``Notations'' paragraph?}\tmcomment{I fixed the inconsistencies. A notations paragraph would be nice, but we don't have space.}
$\f: \bbR^{D_x}\rightarrow \bbR^L$ and $\g: \bbR^{D_y}\rightarrow \bbR^L$ by solving
% \wwcomment{I think we need to say somewhere $\expxy{f(X)^\top g(Y)}$ is computing the expectation of the trace of $f(X)^\top g(Y)$ because both $f$ and $g$ are multi-dimensional?} \tmcomment{ $f(X)^\top g(Y)$ is a scalar by definition. You mean that $f(X)^\top g(Y)=\text{trace}(f(X) g(Y)^\top)$. But we're not using the trace in our derivation anymore, so I don't think we need to mention it.}\wwcomment{You are right. I was mistaken.}
\begin{gather} \label{e:ncca}
  \max_{\f\in\calA,\g\in\calB} \;  \expxy{\f(X)^\top \g(Y)} \\
  \text{s.t.} \;\;  \expx{\f(X)\f(X)^\top}=\expy{\g(Y)\g(Y)^\top}=\I, \nonumber
\end{gather}
where $\calA$ and $\calB$ are two families of (possibly nonlinear) measurable functions.
% (KL)
Observe that if $(\f(\x),\g(\y))$ is a solution to \eqref{e:ncca}, then $(\RR \f(\x),\RR \g(\y))$ is also a solution, for any orthogonal matrix $\RR$.
This ambiguity can be removed by adding the additional constraints $\expxy{f_i(X)g_j(Y)}=0$, $\forall i\neq j$ (see, \eg \citet{Hardoon_04a}). Here we do not pursue this route, and simply focus on one solution among this family of solutions.

\paragraph{Alternating conditional expectations (ACE):} The ACE method \citep{Breiman85} treats the case of a single projection ($L=1$), where $\calB$ is the class of all zero-mean scalar-valued functions $g(Y)$, and $\calA$ is the class of additive models $f(X)=\sum_{\ell=1}^{D_x} \gamma_\ell \phi_\ell(X_\ell)$ with zero-mean scalar-valued functions $\phi_\ell(X_\ell)$.
%\wwcomment{This (sparse) additive model is specific to the \citet{Balakr_12a} paper, right? ACE seems to be a general approach that can be applied to general choice of functions.} \tmcomment{No, this is the general form of ACE. \citet{Balakr_12a} add the sparsity requirement that only a few of the $\gamma_\ell$'s are nonzero}
The ACE algorithm minimizes the objective~\eqref{e:ncca} by iteratively computing the conditional expectation of each view given the other. %This approach, however, does not seem to easily extend to multiple projections ($L>1$), mainly because of the orthogonality constraints in~\eqref{e:ncca}.
Recently, \citet{Makur_15a} extended ACE to multiple dimensions by whitening the vector-valued $\f(X)$ and $\g(Y)$ during each iteration.
%\kl{expand a bit here?}
In practice, the conditional expectations are estimated from training data using nonparametric regression. Since this computationally demanding step has to be repeatedly applied until convergence, ACE and its extensions are impractical to apply on large data.

\paragraph{Kernel CCA (KCCA):} In KCCA \citep{LaiFyfe00a,Akaho01a,Melzer_01a,BachJordan02a,Hardoon_04a}, $\calA$ and $\calB$ are two reproducing kernel Hilbert spaces (RKHSs) associated with user-specified kernels $k_x(\cdot,\cdot)$ and $k_y(\cdot,\cdot)$.  By the representer theorem, the projections can be written in terms of the training samples as $f_\ell(\x)=\sum_{i=1}^N \alpha_{i,\ell} k_x(\x,\x_i)$ and $g_\ell(\y)=\sum_{i=1}^N \beta_{i,\ell} k_x(\y,\y_i)$ with some coefficients $\{\alpha_{i,\ell}\}$ and $\{\beta_{i,\ell}\}$. Letting $\K_x=[k_x(\x_i,\x_j) ]$ and $\K_y=[ k_y(\y_i,\y_j) ]$ denote the $N\times N$ kernel matrices,
% constructed on the training set,
the optimal coefficients can be computed from the top $L$ eigenvectors of the matrix
%$N$-dimensional coefficient vectors $\balpha_\ell=[\alpha_{1,\ell},\dots,\alpha_{N,\ell}]^\top$ and $\bbeta_\ell=[\beta_{1,\ell},\dots,\beta_{N,\ell}]^\top$, can be obtained by computing the top $L$ eigenvectors of the matrix
%\begin{equation}\label{e:kcca}
$(\K_{x} + r_x \I)^{-1} \K_{y} (\K_{y} + r_y \I)^{-1} \K_x$,
%\end{equation}
where $r_x$ and $r_y$ are positive regularization parameters.
%\klcomment{It is a bit odd to throw regularization into the solution without regularizing the objective.}
% that are necessary to avoid the trivial solution.
Computation of the exact solution is intractable for large datasets due to the memory cost of storing the kernel matrices and the time complexity of solving dense eigenvalue systems.  Several approximate techniques have been proposed, largely based on low-rank kernel matrix approximations \citep{BachJordan02a,Hardoon_04a,AroraLivesc12a,Lopez_14b}.  % Weiran: removed citation  WilliamSeeger01a.
%, including the Nystr{\"o}m method \citet{WilliamSeeger01a}, incomplete Cholesky decomposition \citet{BachJordan02a}, partial Gram-Schmidt \citet{Hardoon_04a}, incremental SVD \citet{AroraLivesc12a}, and random Fourier features \citet{Lopez_14b}.
%Most of these methods are based on low-rank approximations to the kernel matrices. %Despite the computational difficulties, KCCA is widely used in various domains \citet{SocherLi10a,AroraLivesc12a,Hodosh_13a}.
\paragraph{Deep CCA (DCCA):} In the more recently proposed DCCA approach \citep{Andrew_13a}, $\calA$ and $\calB$ are the families of functions that can be implemented using two deep neural networks of predefined architecture. As a parametric method, DCCA scales better than approximate KCCA for large datasets \citep{Wang_15b}.

\paragraph{Population solutions:}
\citet{lancaster1958} studied a variant of problem \eqref{e:ncca}, where $\calA$ and $\calB$ are the families of \emph{all} measurable functions. This setting may seem too unrestrictive. However, it turns out that in the population setting, the optimal projections are well-defined even without imposing smoothness in any way. Lancaster characterized the optimal (possibly nonlinear) mappings $f_i$ and $g_i$ for one-dimensional $X$ and $Y$ ($D_x=D_y=1$). In particular, he showed that if $X,Y$ are jointly Gaussian, then the optimal projections are Hermite polynomials. \citet{eagleson1964polynomial} extended this analysis to the Gamma, Poisson, binomial, negative binomial, and hypergeometric distributions. \citet{hannan1961} gave Lancaster's characterization a functional analysis interpretation, which confirmed its validity also for multi-dimensional views.

\paragraph{Our approach:}
Lancaster's population solution has never been used for devising a practical CCA algorithm that works with sample data. Here, we revisit Lancaster's result, extend it to a semi-parametric setting, and devise practical algorithms that work with sample data. Clearly, in the finite-sample setting, it is necessary to impose smoothness. Our approach to imposing smoothness is different from KCCA, which formulates the problem as one of finding the optimal smooth solution (in an RKHS) and then approximates it from samples. Here, we first derive the optimal solution among all (not necessarily smooth) measurable functions, and then approximate it by using smoothed versions of the true densities, which we estimate from data. As we show, the resulting algorithm has significant advantages over KCCA.
\section{Nonparametric and partially linear CCA}
\label{sec:algorithm}
%Rather than working with feature mappings from a user-chosen RKHS or in the form of neural networks, our approach is to derive closed form solutions to the nonlinear CCA problem~\eqref{e:ncca}. We specifically focus on \emph{partially linear CCA} (PLCCA), in which $\calA$ is the set of all linear functions $f(X)=\W^T X$, and $\calB$ is the set of all (nonparametric) functions $g(Y)$, and on \emph{nonparametric CCA} in which both $\calA$ and $\calB$ are the sets of all (nonparametric) functions. In contrast to the ACE algorithm, we address the design of an arbitrary number $L$ of projections. We carry out all derivations in the population setting (assuming $p(\x,\y)$ is known). In practice, we estimate $p(\x,\y)$ from the data.
%Rather than working with feature mappings from a user-chosen RKHS or in the form of neural networks,
%Here, we revisit Lancaster's result, extend it to a semi-parametric setting, and devise practical algorithms which work with sample data.
We treat the following two variants of the nonlinear CCA problem~\eqref{e:ncca}: \ (i)~\emph{Nonparametric CCA} in which both $\calA$ and $\calB$ are the sets of all (nonparametric) measurable functions; \ (ii)~\emph{Partially linear CCA} (PLCCA), in which $\calA$ is the set of all linear functions $\f(\x)=\W^T \x$, and $\calB$ is the set of all (nonparametric) measurable functions $\g(\y)$.
% (KL) removed next sentence since it's already stated elsewhere
%Our projections are not restricted to lie in user-chosen RKHSs (as in KCCA) or to be one-dimensional (as in ACE).
We start by deriving
% (KL)
%closed form
closed-form solutions in the population setting, and then plug in an empirical estimate of $p(\x,\y)$. \kl{reworded}

\subsection{Nonparametric CCA (NCCA)}
\label{sec:NCCA}

Let $\calA$ and $\calB$ be the sets of all (nonparametric) measurable functions of $X$ and $Y$, respectively. Note that the coordinates of $\f(\x)$ and $\g(\y)$ are constrained to satisfy $\expxi{f_i^2(X)}=\expyi{g_i^2(Y)}=1$, so that we may write~\eqref{e:ncca} as an optimization problem over the Hilbert spaces
\begin{align}
\calH_x &= \left\{\left.q:\bbR^{D_x}\rightarrow\bbR \;\right|\;\expxi{q^2(X)}<\infty\right\}, \nonumber\\
 \calH_y &= \left\{\left. u:\bbR^{D_y}\rightarrow\bbR \; \right| \; \expxi{u^2(Y)}<\infty\right\}, \nonumber
\end{align}
which are endowed with the inner products $\langle q, r \rangle_{\calH_x}=\expxi{q(X)r(X)}$ and $\langle u,v \rangle_{\calH_y}=\expyi{u(Y)v(Y)}$. To do so, we express the correlation between $f_i(X)$ and $g_i(Y)$ as
%\begin{align}
%%\sum_{i=1}^L \expxy{f_i(X)g_i(Y)} =
%\sum_{i=1}^L \iint f_i(\x) g_i(\y) p(\x,\y) d\x d\y &= \sum_{i=1}^L \int f_i(\x) \left(\int g_i(\x) \frac{p(\x,\y)}{p(\x)p(\y)} p(\y) d\y\right) p(\x) d\x \nonumber\\
%&= \sum_{i=1}^L \int f_i(\x) \left(\int g_i(\y) S(\x,\y) p(\y) d\y\right) p(\x) d\x \nonumber\\
%&= \sum_{i=1}^L \langle f_i, \SS g_i\rangle_{\calH_x},
%\end{align}
\begin{align}
\expxy{f_i(X)g_i(Y)}
%\sum_{i=1}^L &\!\iint \!\! f_i(\x) g_i(\y) p(\x,\y) d\x d\y \nonumber\\
= \int \!\!f_i(\x)\! \left(\int \!\!g_i(\y) s(\x,\y) p(\y) d\y\!\!\right)\! p(\x) d\x =  \langle f_i, \calS g_i\rangle_{\calH_x},
\end{align}
where\footnote{Formally, $s(\x,\y)$ is the Radon-Nikodym derivative of the joint probability measure w.r.t.\@ the product of marginal measures, assuming the former is absolutely continuous w.r.t.\@ the latter.}
\begin{equation}\label{e:S}
s(\x,\y)=\frac{p(\x,\y)}{p(\x)p(\y)}
\end{equation}
and $\calS:\calH_y\rightarrow\calH_x$ is the operator defined by\footnote{To see that $\calS u\in\calH_x$ for every $u\in\calH_y$, note that $(\calS u)(\x) = \expyi{u(Y)|X=\x}$ and thus $\|\calS u\|^2_{\calH_x} = \expxi{(\expyi{u(Y)|X})^2} \leq \expyi{u^2(Y)} = \|u\|^2_{\calH_y}<\infty$.} $(\calS u)(\x)=\int u(\y) s(\x,\y) p(\y) d\y$. Thus, problem~\eqref{e:ncca} can be written as
\begin{align} \label{e:nccaHilbert}
  \max_{\substack{\langle f_i,f_j\rangle_{\calH_x}=\delta_{ij}\\ \langle g_i,g_j\rangle_{\calH_y}=\delta_{ij}}} \;  &\sum_{i=1}^L \langle \calS g_i, f_i\rangle_{\calH_x} ,
\end{align}
%is the exponentiation of the pointwise mutual information between $X$ and $Y$,
%\begin{align}% \label{e:ncca}
%  \max_{f_1, \ldots f_L, g_1\ldots, g_L} \;  &\sum_{i=1}^L \langle f_i, \SS g_i\rangle_{\calH_x} \quad\quad \text{s.t.} \quad\quad  \langle f_i,f_j\rangle_{\calH_x}=\langle g_i,g_j\rangle_{\calH_y}=\delta_{i,j},
%\end{align}
where $\delta_{ij}$ is Kronecker's delta function.

When $\calS$ is a compact operator, the solution to problem \eqref{e:nccaHilbert} can be expressed in terms of its SVD (see \eg \citep[Proposition A.2.8]{bolla2013}). Specifically, in this case $\calS$ possesses a discrete set of singular values $\sigma_1\geq \sigma_2 \geq \dots$ and corresponding left and right singular functions $\psi_i\in\calH_x,\phi_i\in\calH_y$, and the maximal value of the objective in \eqref{e:nccaHilbert} is precisely $\sigma_1+\ldots+\sigma_L$ and is attained with
\begin{equation}\label{e:nccaSol}
f_i(\x) = \psi_i(\x), \quad  g_i(\y) = \phi_i(\y).
\end{equation}
That is, the optimal projections are the singular functions of $\calS$ and the canonical correlations are its singular values: $\expxyi{f_i(X)g_i(Y)}=\sigma_i$.

\begin{figure}%{0.6\linewidth}
\centering
\includegraphics[width=0.5\linewidth,trim=6.3cm 6.2cm 7.5cm 7.1cm, clip]{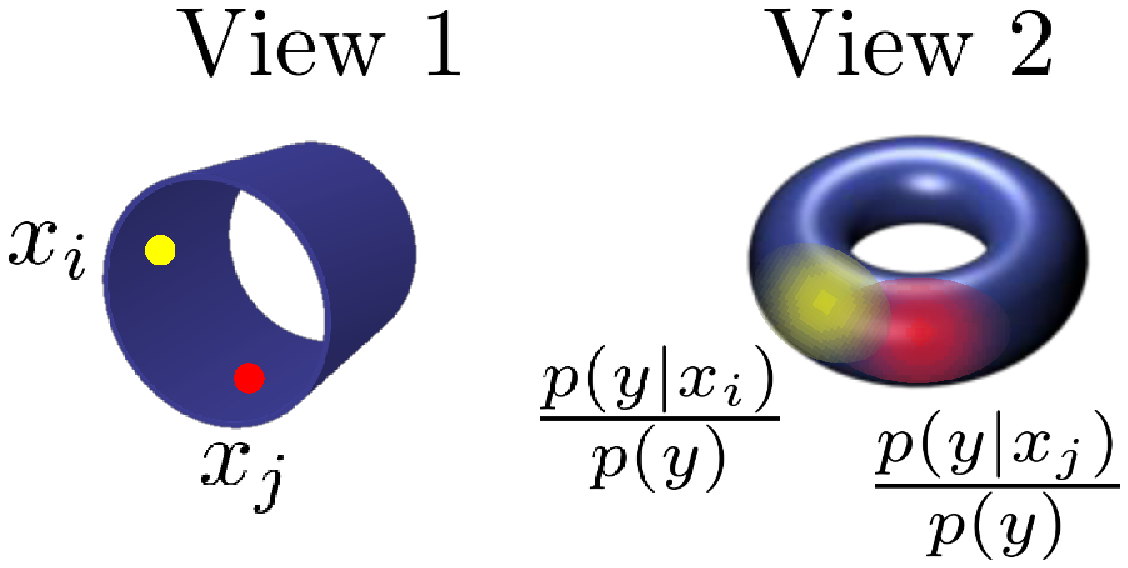}
%\caption{In NCCA, the similarity $k(\x,\x')$ between points $\x$ and $\x'$ in the domain of view 1, is given by the inner product between the functions $p(\y|\x)/p(\y)$ and $p(\y|\x')/p(\y)$ over the domain of view 2.}
\caption{\kl{reworded} In NCCA, the similarity $k(\x,\x')$ between $\x$ and $\x'$ in view 1 is given by the inner product between the functions $p(\y|\x)/p(\y)$ and $p(\y|\x')/p(\y)$ over the domain of view 2.}
\label{f:torus}
\end{figure}

%Before discussing the properties of the NCCA solution~\eqref{e:nccaSol}, let us mention several interesting interpretations.
The NCCA solution~\eqref{e:nccaSol}, has several interesting interpretations. First, note that $\log s(\x,\y)$ is the \emph{pointwise mutual information} (PMI) between $X$ and $Y$, which is a common measure of statistical dependence. Since the optimal projections are the top singular functions of $s(\x,\y)$, the NCCA solution may be interpreted as an embedding which preserves as much of the (exponentiation of the) PMI between $X$ and $Y$ as possible. Second, note that the operator $\calS$ corresponds to the \emph{optimal predictor} (in mean square error sense) of one view based on the other, as $(\calS g_i)(\x)=\expxi{g_i(Y)|X=\x}$ and $(\calS^* f_i)(\y)=\expxi{f_i(X)|Y=\y}$. Therefore, the NCCA projections can also be thought of as approximating the best predictors of each view based on the other. Finally, note that rather than using SVD, the NCCA solution can be also expressed in terms of the \emph{eigen-decomposition} of a certain operator. Specifically, the optimal view $1$ projections are the eigenfunctions of $\calK=\calS\calS^*$ (and the view $2$ projections are eigenfunctions of $\calS^*\calS$), which is the operator defined by $(\calK q)(\x) = \int q(\x) k(\x,\x') p(\x) d\x$, with the kernel
%\begin{align}\label{e:K-def}
%  k(\x,\x')=\expy{s(\x,Y)s(\x',Y)}=\expy{ \left(\frac{p(\x,Y)}{p(\x)p(Y)}\right) \left( \frac{p(\x',Y)}{p(\x')p(Y)} \right) }.
%\end{align}
\begin{align}\label{e:K-def}
  k(\x,\x')=\int s(\x,\y)s(\x',\y)p(\y)d\y.
\end{align}
This shows that NCCA resembles other spectral dimensionality reduction algorithms, in that the projections are the eigenfunctions of some kernel. However, in NCCA, the kernel is not specified by the user. From~\eqref{e:K-def}, we see that $k(\x,\x')$ corresponds to the inner product between $s(\x,\cdot)$ and $s(\x',\cdot)$ \kl{reworded} (equivalently $p(\y|\x)/p(\y)$ and $p(\y|\x')/p(\y)$). Therefore, as visualized in Fig.~\ref{f:torus}, in NCCA $\x$ is considered similar to $\x'$ if the conditional distribution of $Y$ given $X=\x$ is similar to that of $Y$ given $X=\x'$.

%\begin{wrapfigure}{R}{0.6\linewidth}
%\end{wrapfigure}

A sufficient condition for $\calS$ to be compact is that it be a Hilbert-Schmidt operator, \ie that 
\begin{gather*}
\iint |s(x,y)|^2 p(x)dx \,p(y)dy <\infty.
\end{gather*}
Substituting~\eqref{e:S}, this condition can be equivalently written as $\expxyi{s(X,Y)}<\infty$. %%\wwcomment{Just to double-check. Is this a sufficient condition or sufficient and necessary? In the NIPS submission the statement seems to be ``sufficient''.} \tmcomment{Sufficient.}
This can be thought of as a requirement that the statistical dependence between $X$ and $Y$ should not be too strong. In this case, the singular values $\sigma_i$ tend to zero as $i$ tends to $\infty$. Furthermore, the largest singular value of $\calS$ is always $\sigma_1=1$ and is associated with the constant functions $\psi_1(\x)=\phi_1(\y)=1$. To see this, note that for any pair of unit-norm functions $\psi\in\calH_x,\phi\in\calH_y$, we have that $\langle \psi,\calS \phi\rangle_{\calH_x}\!=\!\expxyi{\psi(X)\phi(Y)}\!\leq\!\sqrt{\expxi{\psi^2(X)}\expyi{\phi^2(Y)}}\!=\!1$ and this bound is clearly attained with $\psi(\x)=\phi(\y)=1$. Thus, we see that the first nonlinear CCA projections are always constant functions $f_1(\x)=g_1(\y)=1$. These projections are perfectly correlated, but carry no useful information on the common variability in $X$ and $Y$. Therefore, in practice, we discard them. The rest of the projections are orthogonal to the first and therefore have zero mean:
%\kl{perhaps we should point this out at the outset, and note we could alternatively have a 0 mean constraint}
% (KL)
% so that
 $\expxi{f_\ell(X)}=\expyi{g_\ell(Y)}=0$ for $\ell\geq2$.

\subsection{Partially linear CCA (PLCCA)}
\label{sec:PLCCA}
The above derivation of NCCA can be easily adapted to cases in which $\calA$ and $\calB$ are different families of functions. As an example, we next derive PLCCA, in which $\calA$ is the set of all \emph{linear} functions of $X$ while $\calB$ is still the set of all (nonparametric) measurable functions of $Y$.

Let $\f(\x)=\W^\top \x$, where $\W\in \bbR^{D_x\times L}$. In this case, the constraint that $\expxi{\f(X)\f(X)^\top}=\I$ corresponds to the restriction that $\W^\top\bSigma_{xx}\W=\I$. By changing variables to $\tilde{\W}=\bSigma_{xx}^{1/2}\W$ and denoting the $i$th column of $\tilde{\W}$ by $\tilde{\w}_i$, the constraint simplifies to $\tilde{\w}_i^\top \tilde{\w}_j=\delta_{ij}$. Furthermore, we can write the objective \eqref{e:ncca} as
\begin{align}
\sum_{i=1}^L \expxy{\tilde{\w}_i^\top \bSigma_{xx}^{-\frac{1}{2}}X g_i(Y)} = \sum_{i=1}^L \tilde{\w}_i^\top \expy{\bSigma_{xx}^{-\frac{1}{2}}\expx{X|Y} g_i(Y)} = \sum_{i=1}^L \tilde{\w}_i^\top \calS_{\text{PL}} g_i,
\end{align}
%\begin{equation}
%\sum_{i=1}^L \iint\tilde{\w}_i^\top \bSigma_{xx}^{-\frac{1}{2}}\x g_i(\y)p(\x,\y)d\x d\y = \sum_{i=1}^L \tilde{\w}_i^\top \calS_{\text{PL}} g_i,
%\end{equation}
where $\calS_{\text{PL}}:\calH_y\rightarrow\bbR^{D_x}$ is the operator defined by $\calS_{\text{PL}} u = \bSigma_{xx}^{-1/2} \int \expx{X|Y=\y} u(\y) p(\y) d\y$. Therefore, Problem \eqref{e:ncca} now takes the form %\wwcomment{I suggest using $\tilde{\w}_i^\top \tilde{\w}_j=\delta_{ij}$ in place of $\langle\tilde{\w}_i,\tilde{\w}_j\rangle_{\bbR^{D_x}}=\delta_{ij}$, it is consistent with ``the constraint simplifies to ...''} \tmcomment{Done. I also removed the inner product over $\bbR^{D_x}$ from the objective, to be consistent with the constraint.}
\begin{align} \label{e:plccaHilbert}
  \max_{\substack{\tilde{\w}_i^\top \tilde{\w}_j=\delta_{ij}\\ \langle g_i,g_j\rangle_{\calH_y}=\delta_{ij}}} \;  &\sum_{i=1}^L \tilde{\w}_i^\top \calS_{\text{PL}} g_i,
\end{align}
which is very similar to \eqref{e:nccaHilbert}. Note that here the domain of the operator $\calS_{\text{PL}}$ is infinite dimensional (the space $\calH_y$), but its range is finite-dimensional (the Euclidian space $\bbR^{D_x}$). Therefore, $\calS_{\text{PL}}$ is guaranteed to be compact without any restrictions on the joint probability $p(\x,\y)$. The optimal $\tilde{\w}_i$'s are thus the top $L$ singular vectors of $\calS_{\text{PL}}$ and the optimal $g_i$'s are the top $L$ right singular functions of $\calS_{\text{PL}}$.

The PLCCA solution can be expressed in more convenient form by noting that the optimal $\tilde{\w}_i$'s are also the top $L$ eigenvectors of the matrix $\K_{\text{PL}}=\calS_{\text{PL}}\calS_{\text{PL}}^*$, given by
\begin{align}
\K_{\text{PL}} = \expy{\left(\bSigma_{xx}^{-\frac{1}{2}}\expx{X|Y}\right) \left(\bSigma_{xx}^{-\frac{1}{2}}\expx{X|Y}\right)^\top} = \bSigma_{xx}^{-\frac{1}{2}}\bSigma_{\hat{x}\hat{x}}\bSigma_{xx}^{-\frac{1}{2}}.
\end{align}
Here, $\bSigma_{\hat{x}\hat{x}}=\expyi{\expxi{X|Y}\expxi{X|Y}^\top}$ denotes the covariance of $\hat{X}=\expxi{X|Y}$, the optimal predictor of $X$ from $Y$. Denoting the top $L$ eigenvectors of $\K_{\text{PL}}$ by $\U$, and reverting the change of variables, we get that
% (KL)
%equal to
$\W=\bSigma_{xx}^{-1/2}\U$.

Having determined the optimal $\f(\x)=\W^\top \x$, we can compute the optimal $\g(\y)$ using the following lemma\footnote{A simpler version of this lemma, in which $\f(\x)=\y$ and $\g$ is linear, appeared in~\cite{Eldar_03}. The proof of Lemma~\ref{lem:gy_given_fx} is provided in the Supplemntary Material and follows closely that of \citep[Theorem~1]{Eldar_03}.}.
% \begin{lemma}\label{lem:gy_given_fx}
%   The function $\g$
% % (KL) minimizing
% optimizing \eqref{e:ncca} for a fixed $\f$ is given by %\wwcomment{$f\rightarrow \f$?}
%   \begin{align}\label{e:opt-g}
%     \g(Y)=\left( \expy{ \expxi{\f(X)|Y} \expxi{\f(X)|Y}^\top } \right)^{-\frac{1}{2}} \expxi{\f(X)|Y},
%   \end{align}
%   assuming that the matrix $\expyi{\expxi{\f(X)|Y} \expxi{\f(X)|Y}^\top}$ is non-singular.
% \end{lemma}
\begin{lemma}\label{lem:gy_given_fx}
Assume that $\expyi{\expxi{\f(X)|Y} \expxi{\f(X)|Y}^\top}$ is a non-singular matrix. Then the function $\g$
% (KL) minimizing
optimizing \eqref{e:ncca} for a fixed $\f$ is given by %\wwcomment{$f\rightarrow \f$?}
  \begin{align}\label{e:opt-g}
    \g(Y)=\left( \expy{ \expxi{\f(X)|Y} \expxi{\f(X)|Y}^\top } \right)^{-\frac{1}{2}} \expxi{\f(X)|Y}.
  \end{align}
\end{lemma}
Substituting $\f(\x)=\W^\top \x=\U^\top \bSigma_{xx}^{-1/2} \x$ into \eqref{e:opt-g}, we obtain that the %$g(Y)=(\U^\top \bSigma_{xx}^{-1/2}\bSigma_{xx}\bSigma_{xx}^{-1/2}\U)^{1/2}\U^\top \bSigma_{xx}^{-1/2}\hat{X}$
% (KL)
%partially-linear
partially linear CCA projections are %given by %\wwcomment{$g\rightarrow \g$?}
\begin{align}\label{e:plcca}
\W^\top X = \U^\top \bSigma_{xx}^{-\frac{1}{2}} X, \quad \g(Y) = \D^{-\frac{1}{2}}\U^\top \bSigma_{xx}^{-\frac{1}{2}} \hat{X},
\end{align}
where $\D$ is the diagonal $L\times L$ matrix
% (KL)
%which
that has the top $L$ eigenvalues of $\K_{\text{PL}}$ on its diagonal.

Comparing \eqref{e:plcca} with \eqref{e:ccaL2}, we see that PLCCA has the exact same form as CCA. The only difference is that here $\hat{X}$ is the optimal \emph{nonlinear predictor} of $X$ from $Y$ (a nonlinear function of $Y$), whereas in CCA, $\hat{X}$ corresponded to the best linear predictor of $X$ from $Y$ (a linear function of $Y$).

\subsection{Practical implementations}
%In the previous sections, we presented closed-form solutions to the PLCCA and NCCA problems. These solutions involve moments, conditional expectations, and probability densities which are often not known in practical applications. However, given a set of training data $\{(\x_i,\y_i)\}_{i=1}^N$ drawn independently from the joint density $p(\x,\y)$, these quantities can be estimated and plugged-into our formulas. Different estimation techniques may lead to different algorithms. For concreteness, we now present the algorithms resulting from using the kernel density estimates (KDE)
The NCCA and PLCCA solutions require knowing the joint probability density $p(\x,\y)$ of the views. Given a set of training data $\{(\x_i,\y_i)\}_{i=1}^N$ drawn independently from $p(\x,\y)$, we can estimate $p(\x,\y)$ and plug it into our formulas.  There are many ways of estimating this density.  We next present the algorithms resulting from using one particular choice, namely the kernel density estimates (KDEs)
\begin{align}\label{e:kde}%\textstyle
  &\hat{p}(\x)=\tfrac{1}{N}\sum_{i=1}^N\nolimits w\!\left( \|\x-\x_i\|^2/\sigma_x^2\right), \\ &\hat{p}(\y)=\tfrac{1}{N}\sum_{i=1}^N\nolimits w\!\left( \|\y-\y_i\|^2/\sigma_y^2 \right), \nonumber\\
  &\hat{p}(\x,\y)=\tfrac{1}{N}\sum_{i=1}^N\nolimits w\!\left( \|\x-\x_i\|^2/\sigma_x^2 + \|\y-\y_i\|^2/\sigma_y^2 \right),\nonumber
\end{align}
where $w(t)\propto e^{-t/2}$ is the Gaussian kernel, and $\sigma_x$ and $\sigma_y$ are the kernel widths of the two views.

We note that, theoretically, KDEs suffer from the curse of dimensionality, and use of other density estimation methods is certainly possible. However, we make two important observations. First, real-world data sets often have low-dimensional manifold structure, and the KDE accuracy is affected only by the intrinsic dimensionality. As shown in \citep{OzakinGray09a}, if the data lies on an $r$-dimensional manifold, then the KDE converges to the true density at a rate of\footnote{This requires normalizing the KDE differently, but the scaling cancels out in $s(\x,\y)=p(\x,\y)/p(\x)p(\y)$.} $\mathcal{O}(n^{-\frac{4}{r+4}})$. Indeed, KDEs have been shown to work well in practice in relatively high dimensions~\citep{Georges_03a}, as is also confirmed in our experiments. Second, the NCCA algorithm resulting from working with KDEs involves the same Gaussian affinity matrices used in (Gaussian kernel) KCCA. %% Weiran: I removed this because our W are explicitly normalized, which then are different from Kx Ky. ($\W^x,\W^y$ in Alg.~\ref{alg:ncca} are $\K^x,\K^y$ of KCCA).
 Thus, %% Weiran: I removed  "at least".
 intuitively, the amount of smoothness required for obtaining accurate results in high dimensions is similar for NCCA and KCCA.  Nevertheless, NCCA has a clear advantage over KCCA in terms of both performance and computation.

%sample complexity of learning the KDEs depends only on the much smaller intrinsic dimensionality~\citep{OzakinGray09a}.

\paragraph{PLCCA} Using the above KDEs, the conditional expectation $\hat{\x}(\y)=\expxi{X|Y=\y}$ needed for the PLCCA solution~\eqref{e:plcca} reduces to the Nadaraya-Watson nonparametric regression \citep{nadaraya1964,watson1964}
%\wwcomment{Previously we denoted $\hat{X}=\expxi{X|Y}$, can we use the same notation somehow?}
\begin{align}
  \hat{\x}(\y) = \frac{ \sum_{i=1}^N w \left( \|\y-\y_i\|^2/\sigma_y^2 \right) {\displaystyle \x_i} }{ \sum_{i=1}^N w \left( \|\y-\y_i\|^2/\sigma_y^2 \right) }.
\end{align}
%\begin{align}
%  \exy=\expx{\x|\y}=\frac{ \sum_{i=1}^N G \left( \norm{ (\y-\y_i)/\sigma_y}^2 \right) {\displaystyle \x_i} }{ \sum_{i=1}^N G \left( \norm{(\y-\y_i)/\sigma_y}^2 \right) }.
%\end{align}
%\kl{mention here that we restrict the sum to the NNs?}\wwcomment{The classical estimator does not restrict the sum to the NNs. We did mention we use sparse approximations below.}
The population moments $\bSigma_{\hat{x}\hat{x}}=\expyi{\hat{X}\hat{X}^\top}$ and $\bSigma_{xx}=\expyi{XX^\top}$ can then be replaced by the empirical moments of $\{\hat{\x}(\y_i)\}$ and $\{\x_i\}$.
%
%Then $\widehat{\bSigma}=\expy{\exy \exy^\top}\approx \frac{1}{N} \sum_{i=1}^N \x_y(\y_i) \x_y(\y_i)^\top$ and $\bSigma{xx}=\frac{1}{N} \sum_{i=1}^N \x_i \x_i^\top$ are estimated using the empirical estimators.

\paragraph{NCCA}
%%\tmcomment{I removed everything that has to do with the eigen decomposition of $\K$ and remained only with the SVD of $\SS$. Both views are obtained as the left and right singular vectors of $\SS$. In particular, I removed the part about rescaling the rows of $\SS$ so that $\K=\SS\SS^\top$ is doubly stochastic. This procedure is not symmetric w.r.t. the two views (why not scale the columns of $\SS$?). My only concern is whether we can refer to the Nystrom method when we do SVD rather than eigen-decomposition.} \wwcomment{We need to be careful here as you removed also the details of discretization steps and some things may not be clear to the readers.} \tmcomment{The discretization of the operator $\SS$ is still here. That's all we need now.}
The quadratic form $\langle \calS g_i,f_i\rangle_{\calH_x}$ is given by $\expxi{(\calS g_i)(X)f_i(X)}$ and %thus can be 
is approximated by $\tfrac{1}{N}\sum_{\ell=1}^N (\calS g_i)(\x_\ell) f(\x_\ell)$. Furthermore, $(\calS g_i)(\x_\ell)$ is equal to $\expxi{ s(\x_\ell,Y) g_i(Y) }$ and thus can be approximated by $\tfrac{1}{N} \sum_{m=1}^N s(\x_\ell,\y_m) g(\y_m)$, where $s(\x_\ell,\y_m)= \tfrac{p(\x_\ell,\y_m)}{p(\x_\ell)p(\y_m)}$. Therefore, defining the $N\times N$ matrix $\SS=[s(\x_\ell,\y_m)]$, %\wwcomment{Shall we consider not abusing the notations $\SS$ to avoid any confusion, as of now, it appears to be both an operator and a matrix?} \tmcomment{I'm open to suggestions. How would you like to name it?} \wwcomment{How about for operators, $\K\rightarrow \calK$, $\SS\rightarrow \calS$?}
and stacking the projections of the data points into the $N\times 1$ vectors $\f_i=\tfrac{1}{\sqrt{N}}(f_i(\x_1),\dots, f_i(\x_N))^\top$ and $\g_i=\tfrac{1}{\sqrt{N}}(g_i(\y_1),\dots, g_i(\y_N))^\top$, the NCCA objective can be approximated by $\tfrac{1}{N}\sum_{i=1}^L \f_i^\top \SS \g_i$. Similarly, the NCCA constraints become $\f_i^\top\f_j=\g_i^\top\g_j=\delta_{ij}$. This implies that the optimal $\f_i$ and $\g_i$ are the top $L$ singular vectors of $\SS$. Recall that in the continuous formulation, the first pair of singular functions are constant functions. Therefore, in practice, we compute the top $L+1$ singular vectors of $\SS$ and discard the first one. To construct the matrix $\SS$ we use the kernel density estimates \eqref{e:kde} for joint and marginal probability distributions over $(\x,\y)$.

The NCCA implementation, with the specific choice of Gaussian KDEs, is given in Algorithm~\ref{alg:ncca}. If the input dimensionality is too high, we first perform PCA on the inputs for more robust density estimates.
%Although theoretically KDEs suffer from the curse of dimensionality, we note that real-world data sets often have low-dimensional manifold structures, and the sample complexity of learning the KDEs depends only on the much smaller intrinsic dimensionality~\citep{OzakinGray09a}.  Indeed, KDEs have been shown to work well in practice in relatively high dimensions~\citep{Georges_03a}, as will be confirmed in our experiments.
To make our algorithm computationally efficient, we truncate the Gaussian affinities $\W^x_{ij}$ to zero if $\x_i$ is not within the $k$-nearest neighbors of $\x_j$ (similarly for view 2). This leads to a sparse matrix $\SS$, whose SVD can be computed efficiently.

%\wwcomment{Tomer please double-check this paragraph.}
To obtain out-of-sample mapping for a new view~1 test sample $\x$, we use the Nystr{\"o}m method~\citep{WilliamSeeger01a}, which avoids recomputing SVD. Specifically, recall that the view~1 projections are the eigenfunctions of the positive definite kernel $k(\x,\x')$ of \eqref{e:K-def}. Computing this kernel function between $\x$ and the training samples leads to (notice the corresponding view~2 input of $\x$ is not needed)
\begin{align}\label{e:OutOfSample}
  k(\x,\x_i)=%\frac{1}{N}
\sum_{m=1}^N s(\x, \y_m) s(\x_i, \y_m).
\end{align}
Thus, applying the Nystr{\"o}m method, the projections of $\x$ can be approximated as
%\tmcomment{I fixed this equation and added a description in terms of $S$ rather than $k$.}
%\begin{align*}
%\f(\x) = \sum_{i=1}^N k(\x,\x_i) \Lambda^{-2} \f(\x_i),
%\end{align*}
%where $\Lambda$ contains the top singular values $\sigma_2,\dots,\sigma_{L+1}$ of $\SS$ on its diagonal.
%\begin{align*}
%f_i(\x) = \frac{1}{\sigma_{i+1}^2}\sum_{n=1}^N k(\x,\x_n) f_i(\x_n) = \frac{1}{\sigma_{i+1}}\sum_{n=1}^N s(\x,\y_n)g_i(\y_n)%, \qquad i=1,\dots, L
%\end{align*}
\begin{align*}
f_i(\x) = \frac{1}{\sigma_{i}^2}\sum_{n=1}^N k(\x,\x_n) f_i(\x_n) = \frac{1}{\sigma_{i}}\sum_{n=1}^N s(\x,\y_n)g_i(\y_n)%, \qquad i=1,\dots, L
\end{align*}
%\wwcomment{I modified your second $\y_i \rightarrow \y_n$?}
for $i=1,\dots, L+1$, where $\sigma_i$ is the $i$th singular value of $\SS$. The second equality follows from substituting \eqref{e:OutOfSample} and using the fact that $\f_i$ and $\g_i$ are singular vectors of $\SS$. %Similarly, we apply the Nystr{\"o}m method to the kernel $\SS^\top \SS$ to obtain the view 2 out-of-sample mapping.
Note again that since the affinity matrices are sparse, the mappings are computed via fast sparse matrix multiplication. %\wwcomment{Karen added last sentence.}

\begin{algorithm}[t]
% (KL)
%  \caption{Nonparametric CCA}
  \caption{Nonparametric CCA with Gaussian KDE}
  \label{alg:ncca}
  \renewcommand{\algorithmicrequire}{\textbf{Input:}}
  \renewcommand{\algorithmicensure}{\textbf{Output:}}
  \begin{algorithmic}[1]
    \REQUIRE %% Weiran: I removed "Multi-view".
Training data $\{(\x_i,\y_i)\}_{i=1}^N$, test sample $\x$. %in $\bbR^{D_x} \times \bbR^{D_y}$.
    \STATE Construct affinity matrices for each view
    \begin{gather*}\textstyle
      \W^x_{ij} \leftarrow \exp\left\{ - \frac{\norm{\x_i-\x_j}^2}{2\sigma_x^2} \right\}, \,
      \W^y_{ij} \leftarrow \exp\left\{ - \frac{\norm{\y_i-\y_j}^2}{2\sigma_y^2} \right\}.
    \end{gather*}
    \STATE Normalize $\W^x$ to be right stochastic and $\W^y$ to be left stochastic, \ie
    \begin{gather*}\textstyle
      \W^x_{ij} \leftarrow \W^x_{ij} / \sum_{l=1}^N \W^x_{il} , \;\;
      \W^y_{ij} \leftarrow \W^y_{ij} / \sum_{l=1}^N \W^y_{lj} .
    \end{gather*}
    \STATE Form the matrix $\SS \leftarrow \W^x \W^y$.
    %\STATE (Optional) Rescale rows of $\SS$ iteratively such that $\K=\SS \SS^\top$ is doubly stochastic.
    \STATE Compute $\U\in\bbR^{N\times (L+1)},\V\in\bbR^{N\times (L+1)}$, the first $L+1$ left and right singular vectors of $\SS$, %% Weiran: added:
 with corresponding singular values $\sigma_1,\dots,\sigma_{L+1}$.
    \ENSURE At train time, compute the projections $i=1,\ldots,L+1$ of the training samples as %\kl{also give the out of sample extension, i.e. the output should be the functions we use to map a new test point; or, consider U, V to be the output and don't give the training data projections.  As it is this looks like the end goal is embedding the training points.}
%    \begin{gather*}
%      f_i(\x_n) \leftarrow \sqrt{N} \U_{n,i+1}, \quad%[\f^{2}_i, \dots, \f^{L+1}_i]^\top,
%      g_i(\y_n) \leftarrow \sqrt{N} \V_{n,i+1}, \quad i=1,\ldots,L.%[\f^{2}_i, \dots, \f^{L+1}_i]^\top,
%    \end{gather*}
     \begin{gather*}
      f_i(\x_n) \leftarrow \sqrt{N} \U_{n,i}, \quad%[\f^{2}_i, \dots, \f^{L+1}_i]^\top,
      g_i(\y_n) \leftarrow \sqrt{N} \V_{n,i}.%[\f^{2}_i, \dots, \f^{L+1}_i]^\top,
    \end{gather*}
    \tmcomment{I added the out-of-sample extension. Let me know if its clear.}\wwcomment{Modified. Double-check.}
    %% At test time, concatenate a new row to $\W^x$, corresponding to the test sample $\x$, and accordingly update $\SS$ to be $(N+1)\times N$. Compute the projections of $\x$ as
    At test time, calculate a new row of $\W^{x}$ for $\x$ as
\begin{gather*}\textstyle
\W^{x}_{N+1,j} \leftarrow \exp\left\{ - \frac{\norm{\x-\x_j}^2}{2\sigma_x^2} \right\},\\
\textstyle
\W^x_{N+1,j} \leftarrow \W^x_{N+1,j}/\sum_{l=1}^N \W^x_{N+1,l}
\end{gather*}
%% and accordingly update $\SS$ to be $(N+1)\times N$. Compute the projections of $\x$ as
and a new row of $\SS$ as $\SS_{N+1}\leftarrow \W^{x}_{N+1} \W^{y}$, and compute the projections of $\x$ as
%     \begin{gather*}
%       f_i(\x) \leftarrow \frac{1}{\sigma_{i+1}} \sum_{n=1}^N \SS_{N+1,\,n} \,\,g_i(\y_n), \quad i=1,\ldots,L,%[\f^{2}_i, \dots, \f^{L+1}_i]^\top,
%     \end{gather*}
%    \begin{gather*}
%      f_i(\x) \leftarrow \frac{1}{\sigma_{i+1}} \sum_{n=1}^N  \SS_{N+1,n} \,\,g_i(\y_n), \quad i=1,\ldots,L,%[\f^{2}_i, \dots, \f^{L+1}_i]^\top,
%    \end{gather*}
        \begin{gather*}
      f_i(\x) \leftarrow \frac{1}{\sigma_{i}} \sum_{n=1}^N  \SS_{N+1,n} \,\,g_i(\y_n), \quad i=1,\ldots,L+1. %[\f^{2}_i, \dots, \f^{L+1}_i]^\top,
    \end{gather*}
%% Weiran: removed    where $\sigma_i$ is the %$i^\textrm{th}$
%% $i$th singular value of the original ($N\times N$ matrix) $\SS$.
%% $\SS$
    %where $f^{j}_i$ is the $i$-th element of $\f^{j}$, $j=2,\dots,L+1$.
  \end{algorithmic}
\end{algorithm}

\paragraph{Relationship with KCCA}
Notice that NCCA is not equivalent to KCCA with any kernel. KCCA requires two kernels, each of which only sees one view; the NCCA kernel \eqref{e:K-def} depends on both views through their joint distribution. In terms of practical implementation, our KDE-based NCCA solves a different eigenproblem and does not involve any full matrix inverses. Indeed, both methods compute the SVD of the matrix $\Q_x^{-1} \W^x \W^y \Q_y^{-1}$. However, in NCCA, $\Q_x,\Q_y$ are diagonal matrices containing the sums of rows/columns of $\W^x/\W^y$, whereas in KCCA, $\Q_x=\W^x+r_x\I$, $\Q_y=\W^y+r_y\I$, for some positive regularization parameters $r_x,r_y$. Moreover, in NCCA this factorization gives the projections, whereas in KCCA it gives the coefficients in the RKHS.

An additional key distinction is that NCCA does not require regularization in order to be well defined. In contrast, KCCA must use regularization, as otherwise the matrix it factorizes collapses to the identity matrix, and the resulting projections are meaningless. This is due to the fact that KCCA attempts to estimate covariances in the infinite-dimensional feature space, whereas NCCA is based on estimating probability densities in the primal space.

The resulting computational differences are striking.  The number of training samples $N$ is often such that the $N \times N$ matrices in either NCCA or KCCA cannot even be stored in memory.  However, these matrices are sparse, with only $kN$ entries if we retain $k$ neighbors.  Therefore, in NCCA the storage problem is alleviated and matrix multiplication and eigendecomposition are $O(kN^2)$ operations instead of $O(N^3)$.  In KCCA, one cannot take advantage of truncated kernel affinities, because of the need to compute the inverses of kernel matrices, which are in general not sparse, so direct computation is often infeasible in terms of both memory and time.  Low-rank KCCA approximations (as used in our experiments below) with rank $M$ have a time complexity $O(M^3+M^2 N)$, %%\wwcomment{I corrected the complexity, it was $O(M^3+M N^2)$ which was incorrect.}
which is still challenging with typical ranks in the thousands or tens of thousands.
%%\wwcomment{Now that we have the large scale KCCA implementation, I wonder if we shall weaken such statement.} 
\section{Related work}
\label{sec:related}
%The NCCA algorithm is related to several other unsupervised multi-view learning methods.
%The study of the optimal single pair of nonlinear projections of two vectors dates back to R{\'e}nyi \citet{Renyi59}, who presented the solution as the top eigenfunction of a certain operator.%, and gave conditions under which the maximal correlation is attained.
% (guaranteed by the compactness of the operator).
%The study of the optimal \emph{single} pair of nonlinear projections dates back to R{\'e}nyi \citet{Renyi59}, who presented the solution as the top eigenfunction of a certain operator. %, and gave conditions under which the
%NCCA extends R{\'e}nyi's analysis to multiple pairs of projections, by incorporating orthogonality constraints on the learned dimensions. In contrast to \citet{Renyi59}, we also provide a concrete algorithm for approximating the population solution.
% from a training set.

Several recent multi-view learning algorithms use products or sums of single-view affinity matrices, diffusion matrices, or Markov transition matrices.
% (and their transposes).
% In this respect, t
The combined kernels constructed in these methods resemble our matrix $\SS=\W^x \W^y$. Such an approach has been used, for example, for multi-view spectral clustering \citep{Sa05a,
%Sa_10a,
ZhouBurges07a,Kumar_11a}, metric fusion~\citep{Wang_12c}, common manifold learning~\citep{LedermTalmon14a}, and multi-view nonlinear system identification~\citep{BootsGordon12a}.  Note, however, that in NCCA the matrix $\SS$ corresponds to the product $\W^x\W^y$ \emph{only when using a separable Gaussian kernel} for estimating the joint density $p(\x,\y)$. If a non-separable density estimate is used, then the matrix $\SS$ no longer resembles the previously proposed multi-view kernels.
%\kl{so they are special cases of NCCA?}\wwcomment{I do not think so, because of the different normalization used and also the eigensystems are somewhat different for each method.}
Furthermore, although algorithmically similar, NCCA arises from a completely different motivation:  It maximizes the correlation between the views, whereas these other methods do not.

\section{Experiments}
\label{s:expt}

In the following experiments, we compare PLCCA/NCCA with linear CCA, two kernel CCA approximations using random Fourier features (FKCCA, \cite{Lopez_14b}) and Nystr{\"o}m approximation (NKCCA, \cite{WilliamSeeger01a}) as described in \cite{Wang_15b}, and deep CCA (DCCA, \cite{Andrew_13a}). % Weiran: removed "on several tasks."
% the multi-view representation learning setting, where only one view is available at test time and we reply on the features (projection mappings) learned with two-view data during training.

\begin{figure*}[t]
\centering
\begin{tabular}{@{}c@{\hspace*{0.005\linewidth}}c@{\hspace*{0.005\linewidth}}c@{\hspace*{0.005\linewidth}}c@{\hspace*{0.005\linewidth}}c@{\hspace*{0.005\linewidth}}c@{}}
(a) View 1  & (b) View 2 & (c) NKCCA & (d) DCCA & (e) PLCCA & (f) NCCA \\[-.5ex]
  \includegraphics[width=0.16\linewidth,height=0.16\linewidth]{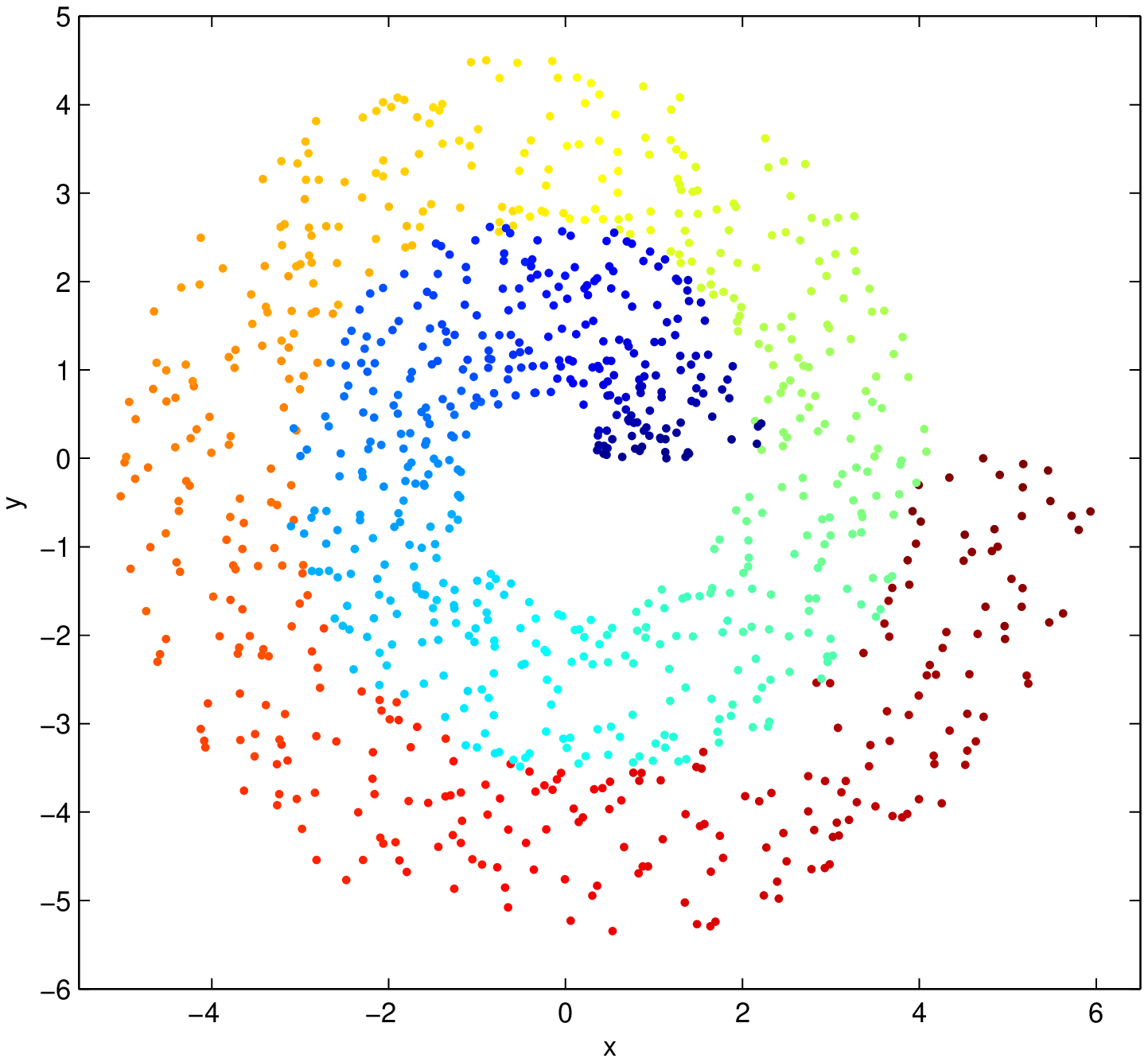} &
  \includegraphics[width=0.16\linewidth,height=0.16\linewidth]{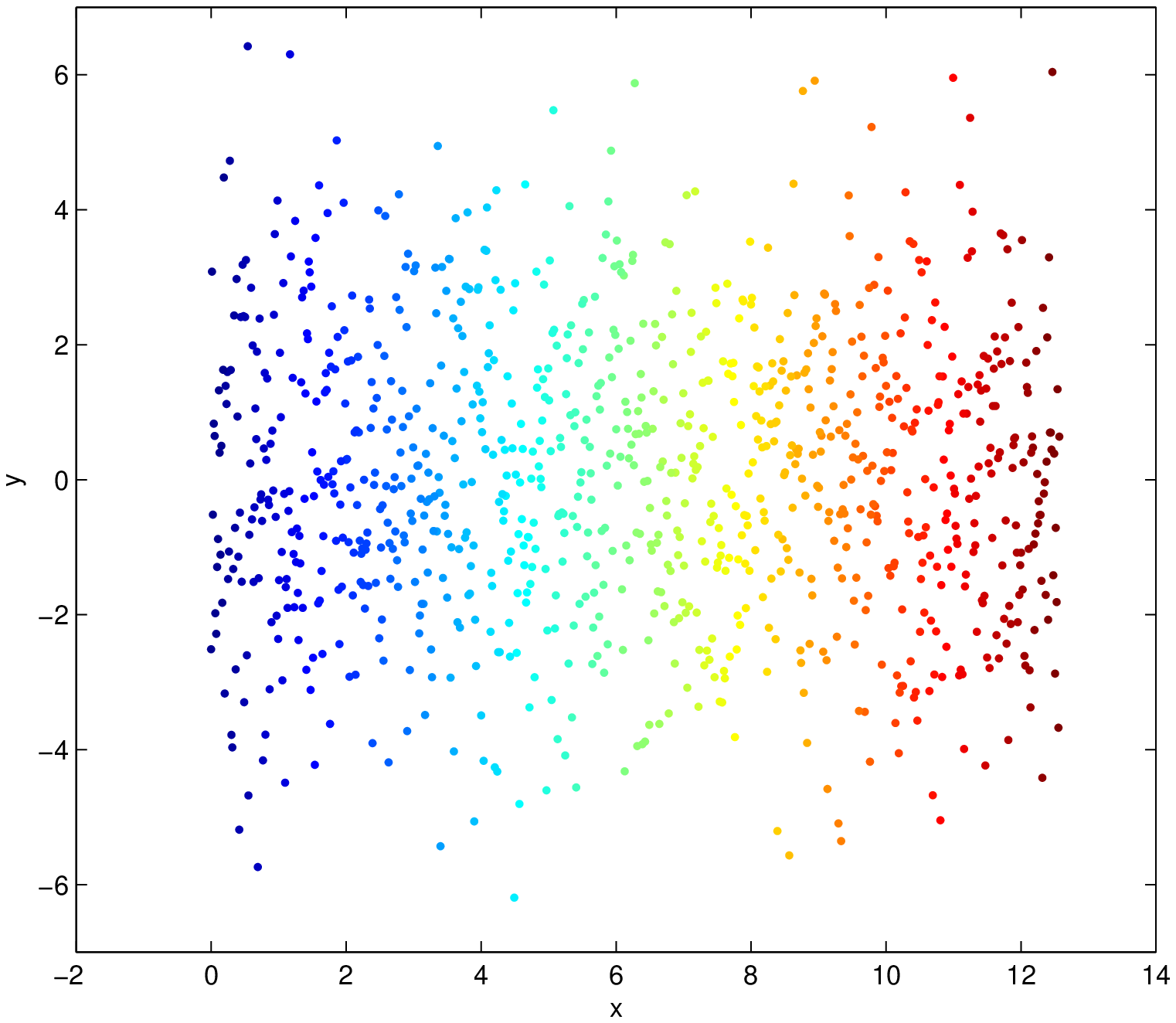} &
\psfrag{g(x)}[t][][0.6]{$f(X)$}
\psfrag{h(y)}[][][0.6]{$g(Y)$}
  \includegraphics[width=0.16\linewidth]{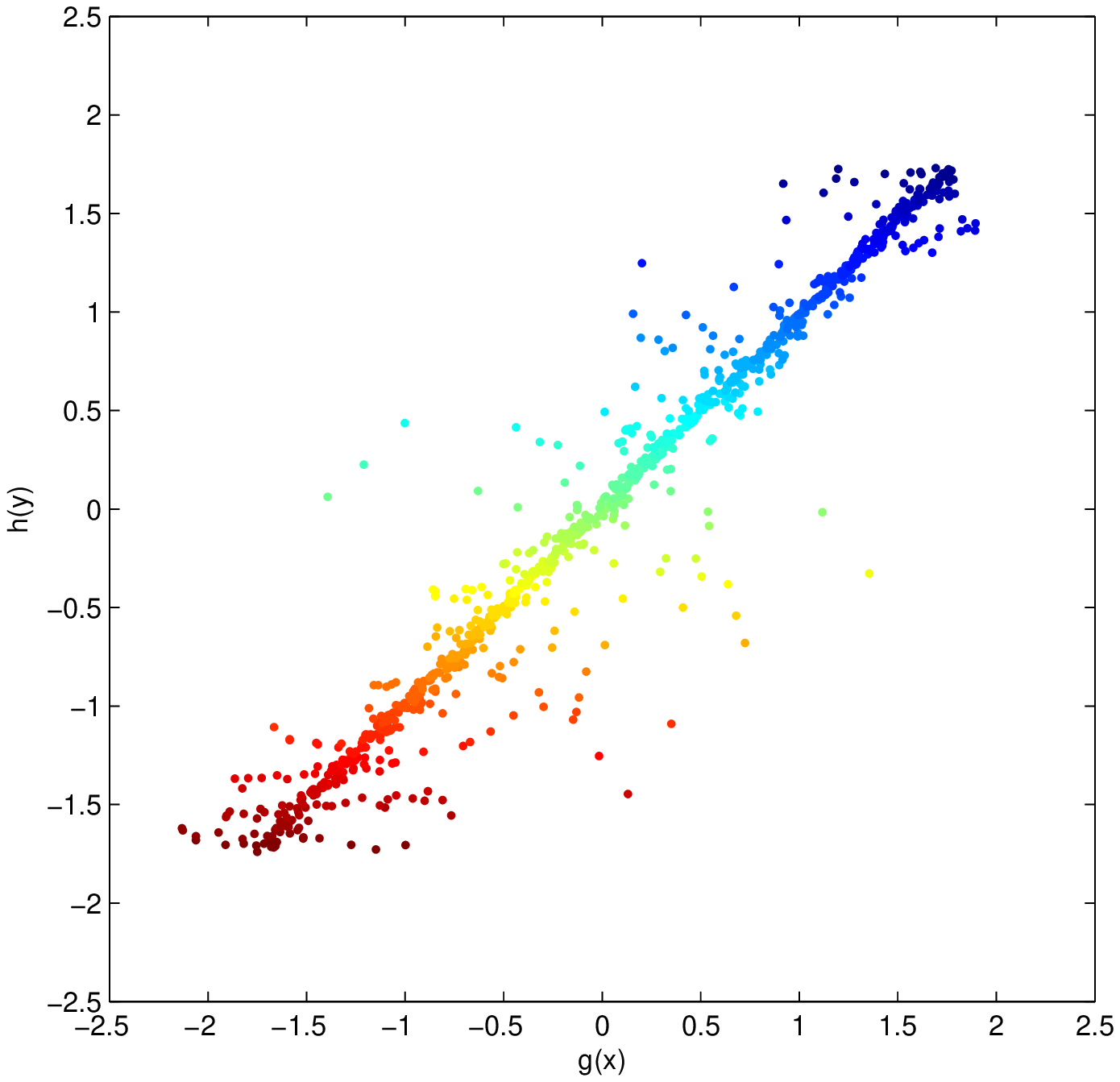} &
\psfrag{g(x)}[][]{}
\psfrag{h(y)}[][]{}
  \includegraphics[width=0.16\linewidth]{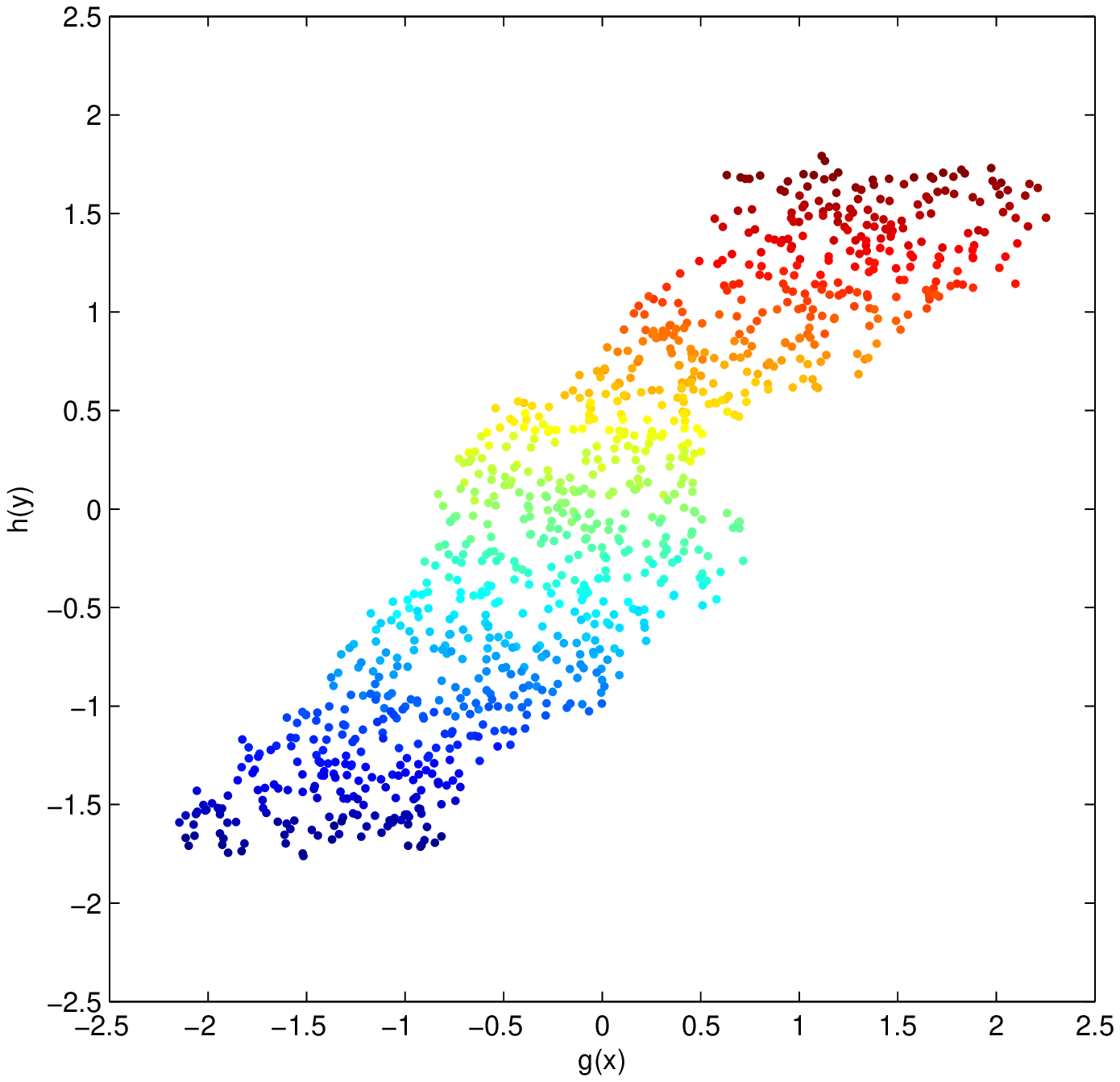} &
\psfrag{g(x)}[][]{}
\psfrag{h(y)}[][]{}
  \includegraphics[width=0.16\linewidth]{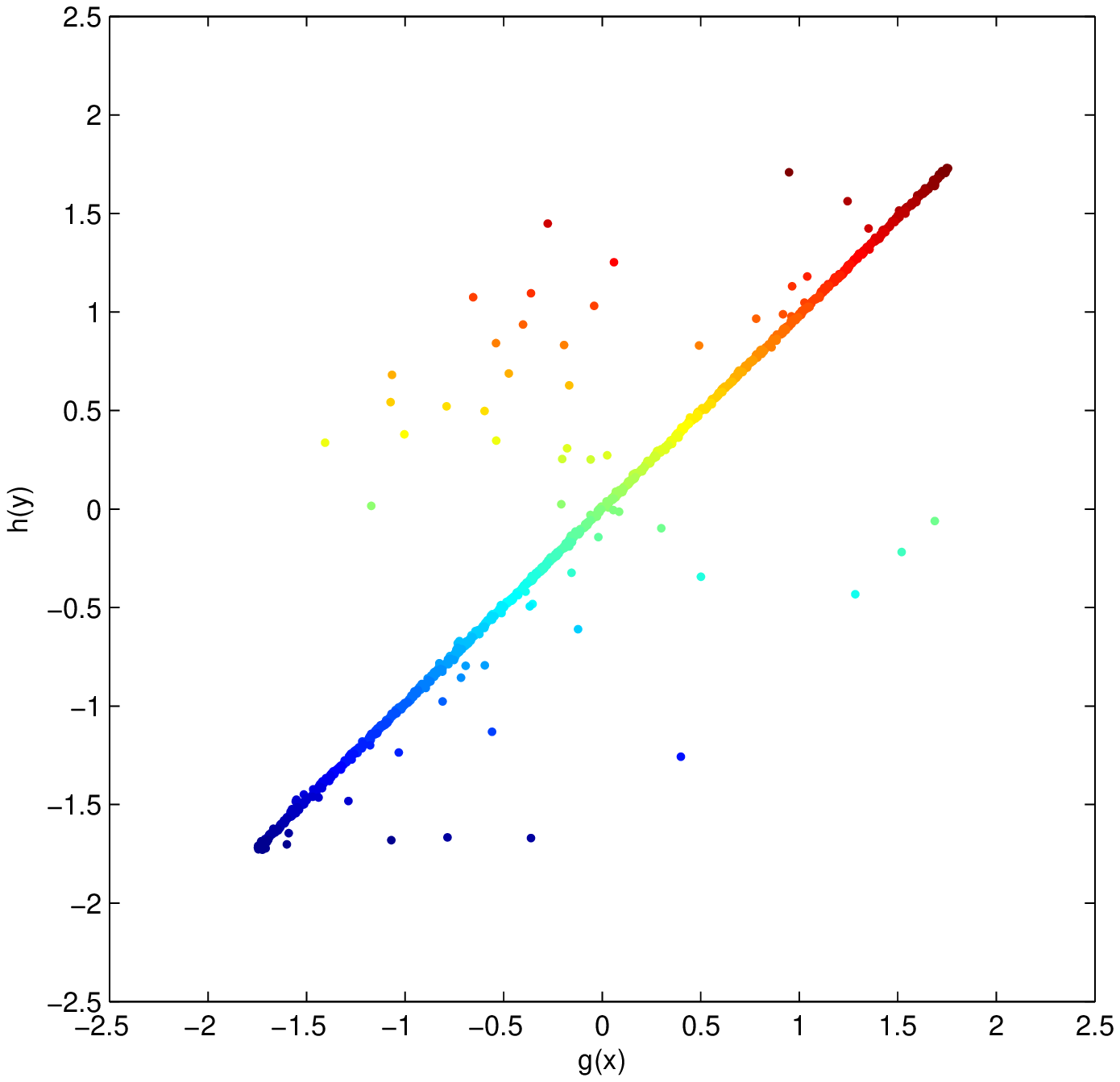} &
\psfrag{g(x)}[][]{}
\psfrag{h(y)}[][]{}
  \includegraphics[width=0.16\linewidth]{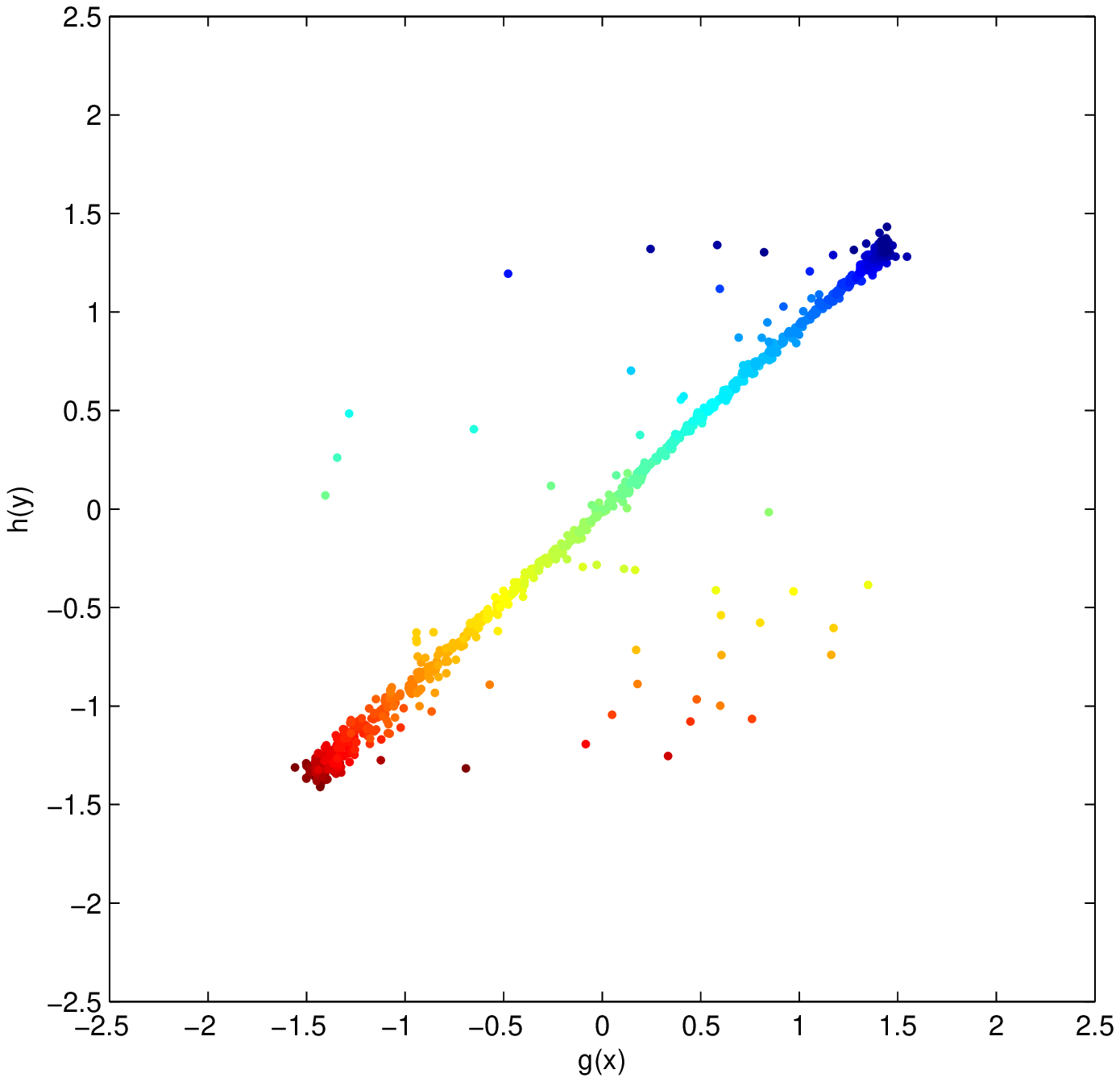}
\end{tabular}
\caption{Dimensionality reduction obtained by nonlinear CCAs on a synthetic dataset.}
\label{f:toy}
\end{figure*}

\paragraph{Illustrative example} We begin with the 2D synthetic dataset ($1000$ training samples) in Fig.~\ref{f:toy}(a,b), where samples of the two input manifolds are colored according to their common degree of freedom. Clearly, a linear mapping in view 1 cannot unfold the manifold to align the two views, and linear CCA indeed fails (results not shown). We extract a one-dimensional projection for each view using different nonlinear CCAs, and plot the projection $g(\y)$ vs.~$f(\x)$  of test data (a different set of $1000$ random samples from the same distribution) in Fig.~\ref{f:toy}(c-f). Since the second view is essentially a linear manifold (plus noise), for NKCCA we use a linear kernel in view 2 and a Gaussian kernel in view~1, and for DCCA we use a linear network for view~2 and two hidden layers of $512$ ReLU units for view~1. Overall, NCCA achieves better alignment of the views while compressing the noise (variations not described by the common degree of freedom). While DCCA also succeeds in unfolding the view~1 manifold, it fails to compress the noise.
% as it is hard for a smooth parametric function to denoise view 1.
%%\kl{this sounds weak -- we could make DCCA less smooth if we want.  It sounds like we didn't tune enough.} \wwcomment{at the time I tried very large number of hidden units, and it was still not able to separate the branches as nicely.} \kl{I removed a phrase}

% \begin{table}[t]
% \centering
% \caption{Total canonical correlation achieved by each algorithm on the XRMB 'JW11' test set.}
% \label{t:jw11_corr}
% \begin{tabular}{|c||c|c|c|c|c|c|}\hline
%  & CCA & FKCCA & NKCCA & NCCA  & DCCA L-BFGS & DCCA Stochastic \\ \hline
% Canon. Corr. & 23.4 &  68.2 & 74.3 & 73.4 & 83.7 & 89.6 \\ \hline
% \end{tabular}
% \end{table}

\begin{table*}[ht]
\centering
% (KL)
\caption{Total canonical correlation on the XRMB 'JW11-s' test set and run time of each algorithm.  The maximum possible canonical correlation is 112 (the view 2 input dimensionality).  PLCCA/NCCA run time is given as neighbor search time + optimization time.}
\label{t:jw11}
%\begin{small}
\begin{tabular}{|c||c|c|c|c|c|c|}\hline
                   & CCA &  FKCCA & NKCCA  & DCCA    & PLCCA & NCCA  \\ \hline
Total Correlation  & 21.7 &  99.2 & 105.6   & 107.6   & 79.4  & 107.9 \\ \hline
Run Time (sec)     & 2.3  & 510.7 & 1449.8 & 10044.0 & 40.7 + 0.8  & 69.4 + 79.0 \\ \hline
%Run time (seconds) & 2.3  & 510.7 & 1449.8 & 10044.0 & 41.5  & 148.4 \\ \hline
\end{tabular}
%\end{small}
\end{table*}
% Exact 15 nearest neighbor search time at 20\% of input dimensionality on JW11 takes 40.7/28.7 seconds for view 1/2.

% \begin{wrapfigure}{R}{0.36\linewidth}
% \centering
% \begin{tabular}{@{}cc@{}}
% \rotatebox{90}{\hspace{3em}PLCCA} &
% \includegraphics[width=0.97\linewidth]{mnist_PLCCA.eps} \\[-1ex]
% \rotatebox{90}{\hspace{4em}NCCA} &
% \includegraphics[width=0.83\linewidth]{mnist_NCCA.eps}
% \end{tabular}
% \caption{2D visualization by t-SNE of PLCCA/NCCA projections of the noisy MNIST test set.}
% \label{f:mnist_emb}
% \end{wrapfigure}

% \begin{table}[t]
% \centering
% \caption{Classification error rates (\%, $\downarrow$) of a linear SVM on the projected test set of noisy MNIST (50K training set).}
% \label{t:mnist}
% \begin{tabular}{|c||c|c|c|c|c|c|c|}\hline
%                      & Baseline & CCA   & FKCCA & NKCCA & DCCA  & PLCCA  & NCCA   \\ \hline
% Error Rates &         13.1      & 19.6  & \hspace*{.5em}5.1  &  \hspace*{.5em}4.5  & \hspace*{.5em}2.9   & \hspace*{.5em}4.4    & \hspace*{.5em}4.1  \\ \hline
% \end{tabular}
% \end{table}

\paragraph{X-Ray Microbeam Speech Data}
The University of Wisconsin X-Ray Micro-Beam (XRMB) corpus \citep{Westbur94a} consists of simultaneously recorded speech and articulatory measurements. Following \citet{Andrew_13a} and \citet{Lopez_14b}, the acoustic view inputs are 39D Mel-frequencey cepstral coefficients and the articulatory view inputs are horizontal/vertical displacement of 8 pellets attached to different parts of the vocal tract, each then concatenated over a 7-frame context window, for speaker 'JW11'. % We use the same split of \citet{Andrew_13a} for speaker 'JW11' where $30K$/$10K$/$11K$ frames % \footnote{Notice that \citet{Lopez_14b} used a different (random) split of the data and so their results are different.} are used for training/tuning/testing.
As in \citep{Lopez_14b}, we randomly shuffle the frames and generate splits of $30K$/$10K$/$11K$ frames for training/tuning/testing, and we refer to the result as the 'JW11-s' setup (random splits better satisfy the i.i.d.~assumption of train/tune/test data than splits by utterances as in \citep{Andrew_13a}). We extract $112D$ projections with each algorithm and measure the total correlation between the two views of the test set, after an additional $112D$ linear CCA.  As in prior work, for both FKCCA and NKCCA we use rank-$6000$ approximations for the kernel matrices; for DCCA we use two ReLU~\citep{NairHinton10a} hidden layers of width $1800$/$1200$ for view 1/2 respectively and run stochastic optimization with minibatch size $750$ as in \citep{Wang_15a} for $100$ epochs. Kernel widths for FKCCA/NKCCA, learning rate and momentum for DCCA, kernel widths and neighborhood sizes for NCCA/PLCCA are selected by grid search based on total tuning set correlation. Sensitivity to their values is mild over a large range; \eg setting the kernel widths to 30-60\% of the sample $L_2$ norm gives similarly good results. For NCCA/PLCCA, input dimensionalities are first reduced by PCA to 20\% of the original ones (except that PLCCA does not apply PCA for view 2 in order to extract a $112D$ projection). The total correlation achieved by each algorithm is given in Table~\ref{t:jw11}. We also report the running time (in seconds) of the algorithms (measured with a single thread on a workstation with a 3.2GHz CPU and 56G main memory), each using its optimal hyperparameters, and including the time for exact $15$-nearest neighbor search for NCCA/PLCCA. Overall, NCCA achieves the best canonical correlation while being much faster than the other nonlinear methods.

\begin{table*}[t]
\centering
% (KL)
\caption{Clustering accuracy, SVM error rate, and run times %%
(same format as in Table~\ref{t:jw11})  on the noisy MNIST projected test set. % PLCCA/NCCA run time is given as kNN search time + optimization time.
}
\label{t:mnist}
%\begin{small}
\begin{tabular}{|c||c|c|c|c|c|c|c|}\hline
                     & Baseline & CCA   & FKCCA & NKCCA & DCCA  & PLCCA  & NCCA   \\ \hline
%Clust. Acc. (\%, $\uparrow$) &         47.1      & 72.3  & 95.6  & 96.7  & 99.1  & 98.4   & 99.2 \\ \hline
Clustering Accuracy (\%) &         47.1      & 72.3  & 95.6  & 96.7  & 99.1  & 98.4   & 99.2 \\ \hline
%Error Rate (\%, $\downarrow$) &         13.3      & 18.9  & \hspace*{.5em}3.9  &  \hspace*{.5em}3.1  & \hspace*{.5em}0.9   & \hspace*{.5em}1.3    & \hspace*{.5em}0.7  \\ \hline
Classification Error (\%) &         13.3      & 18.9  & \hspace*{.5em}3.9  &  \hspace*{.5em}3.1  & \hspace*{.5em}0.9   & \hspace*{.5em}1.3    & \hspace*{.5em}0.7  \\ \hline
Run Time (sec) & 0 & 161.9 & 1270.1 & 5890.3 & 16212.7 & 4932.1 + 5.7 & 9052.6 + 38.3 \\ \hline
\end{tabular}
%\end{small}
\end{table*}
% Time (seconds) & 0 & 161.9 & 1270.1 & 5890.3 & 16212.7 (23 epochs) & 4937.8 & 9090.9
% Included time for exact 5/10 nearest neighbor search time at 100 dimensions on MNIST, which takes 4120.5/4932.1 seconds for view 1/2.

%\begin{wrapfigure}{R}{0.6\linewidth}
\begin{figure}
\centering
\begin{tabular}{@{}c@{\hspace*{0em}}c@{\hspace*{0em}}c@{\hspace*{0em}}c@{}}
NKCCA & DCCA & PLCCA & NCCA\\
\includegraphics[width=0.31\linewidth,bb=95 245 470 560,clip]{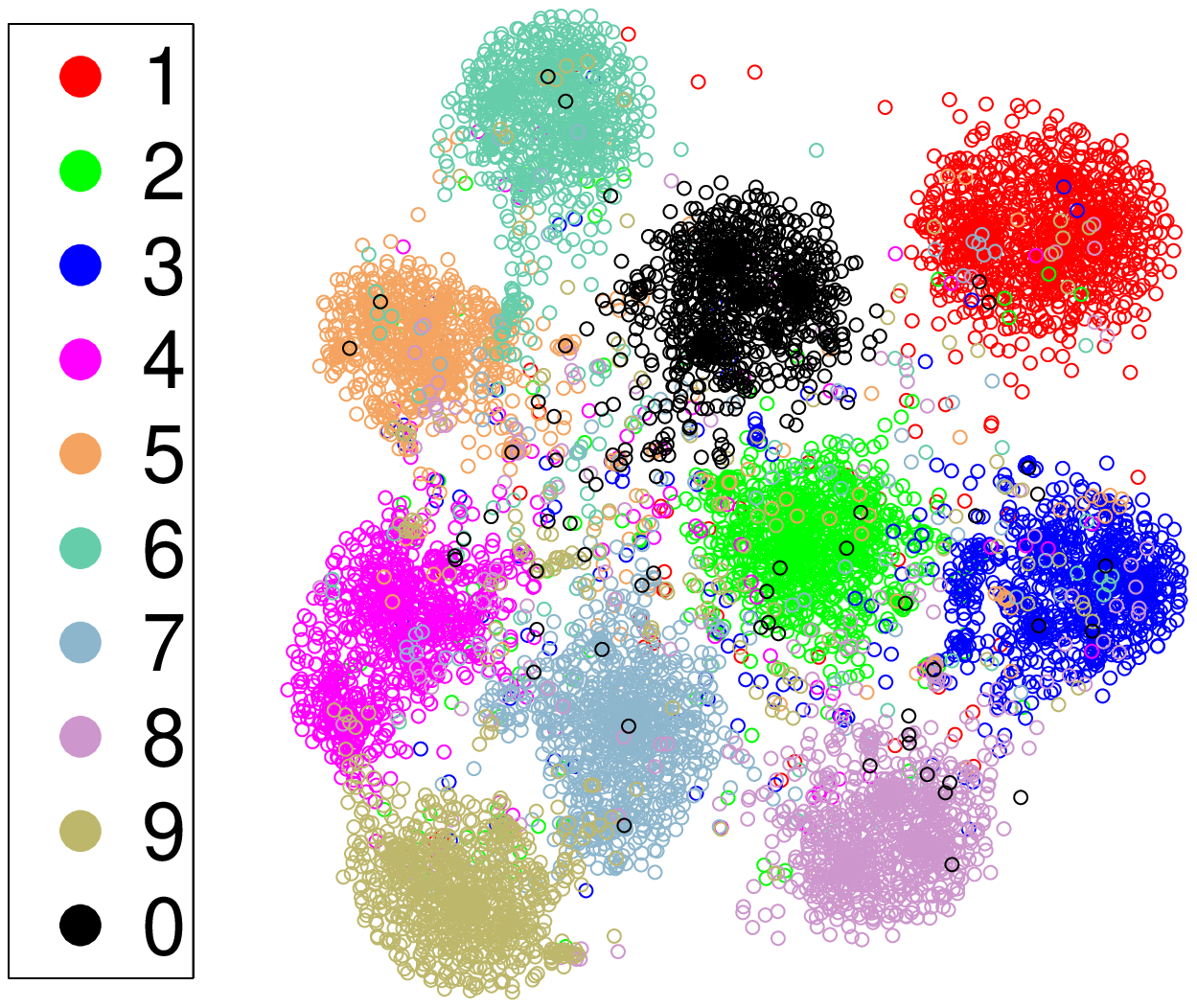} &
\includegraphics[width=0.23\linewidth]{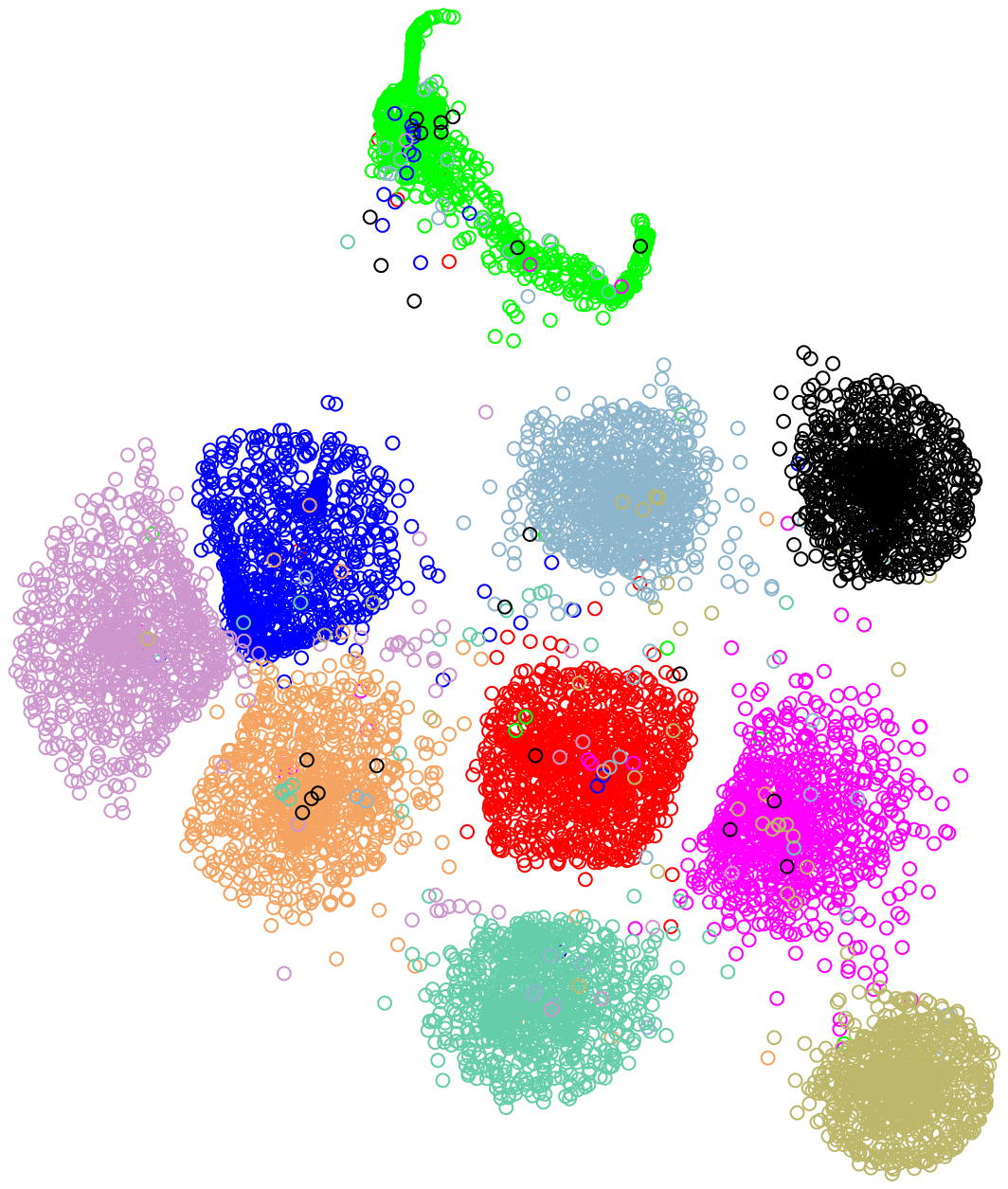} &
\includegraphics[width=0.23\linewidth]{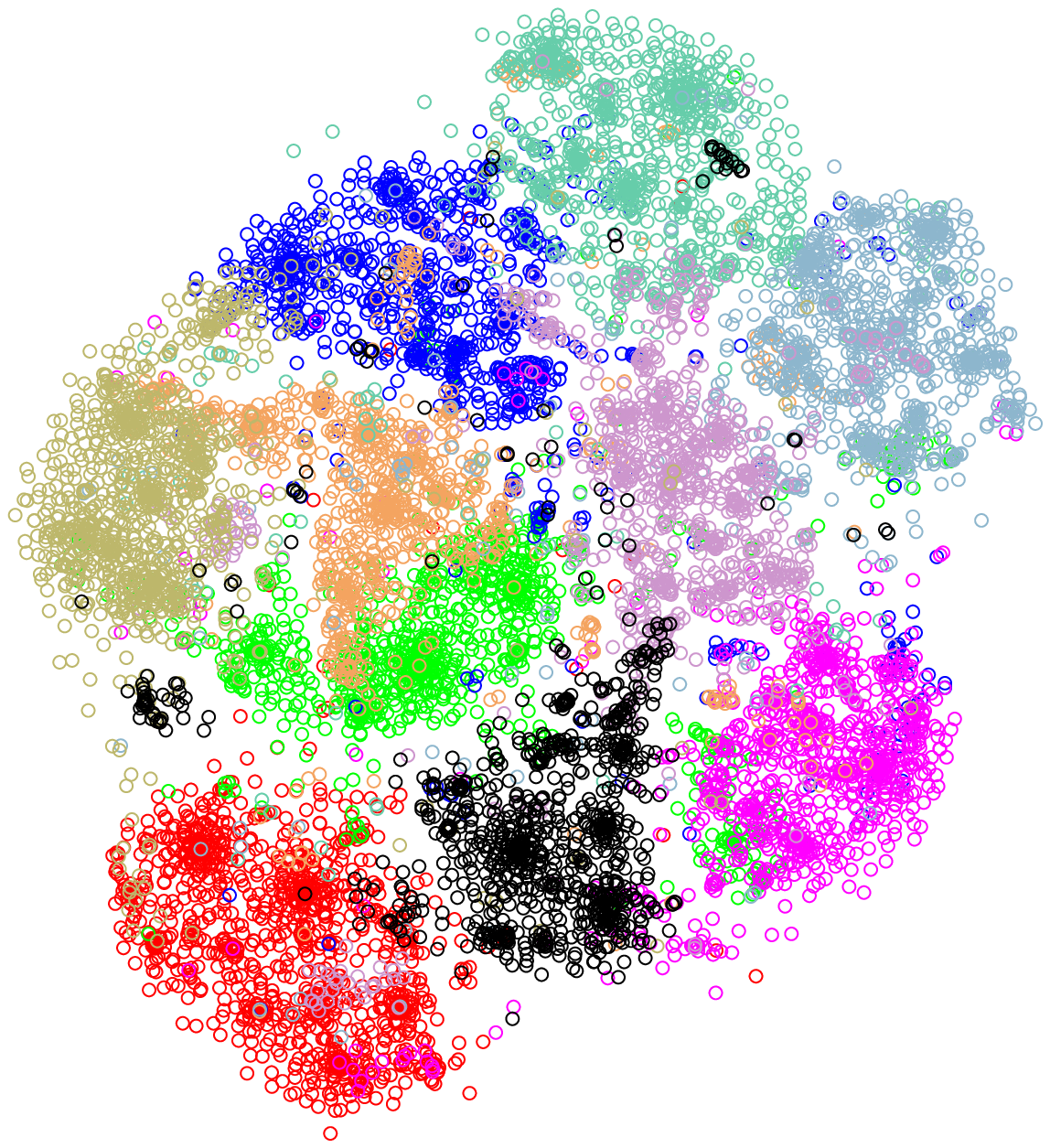} &
\includegraphics[width=0.23\linewidth]{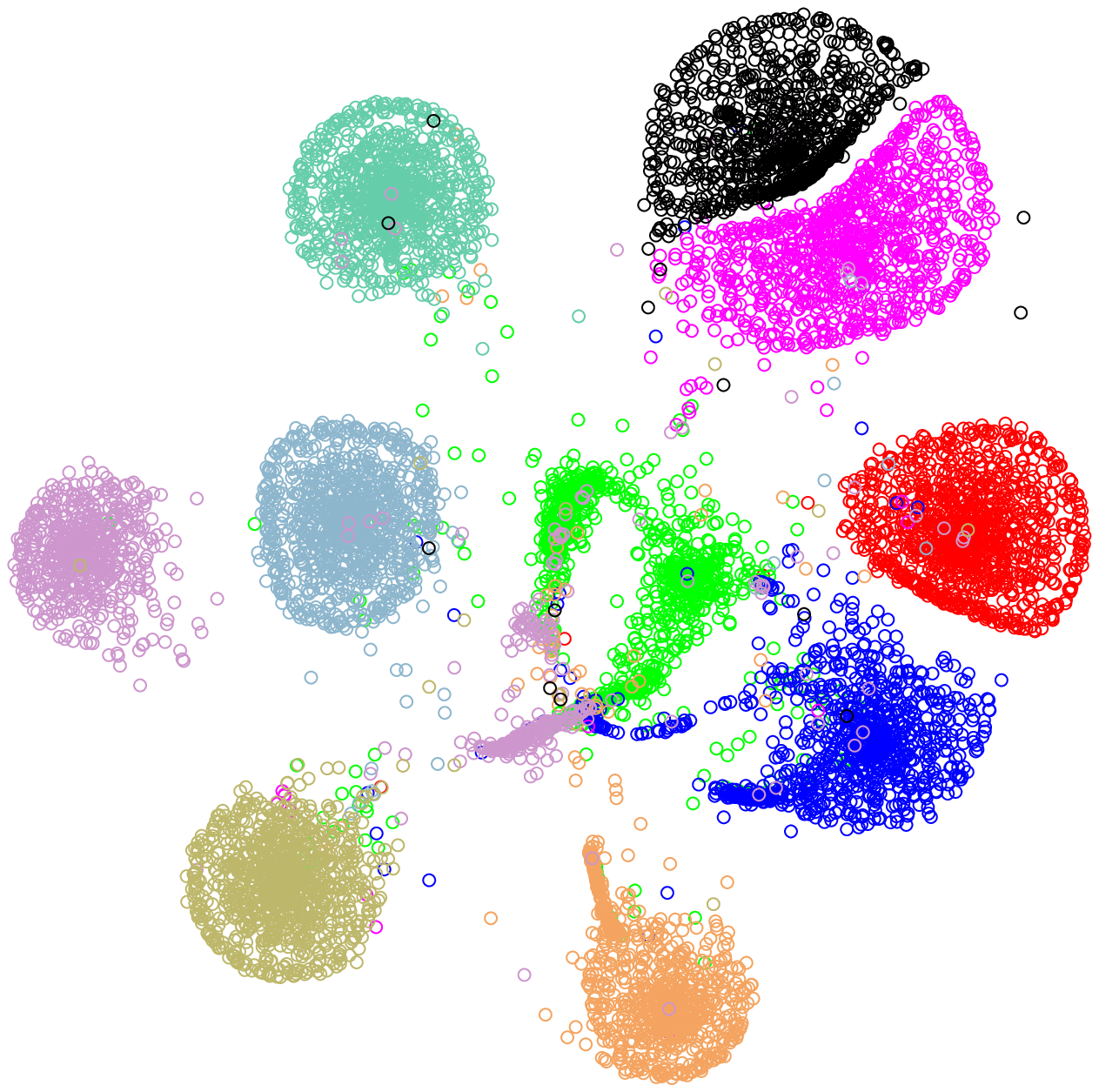}
\end{tabular}
\caption{2D t-SNE visualization of the noisy MNIST test set.}
\label{f:mnist_emb}
\end{figure}
%\end{wrapfigure}

\paragraph{Noisy MNIST handwritten digits dataset} We now demonstrate the algorithms on a noisy MNIST dataset, generated identically to that of \citet{Wang_15b} but with a larger training set.  View~1 inputs are randomly rotated images ($28\times 28$, gray scale) from the original MNIST dataset \citep{Lecun_98a}, and the corresponding view~2 inputs are randomly chosen images with the same identity plus additive uniform pixel noise.  We generate $450K$/$10K$/$10K$ pairs of images for training/tuning/testing (\citet{Wang_15b} uses a $50K$-pair training set).  This dataset satisfies the multi-view assumption
%\citet{Chaudh_09a}, i.e.,
that given the label, the views are uncorrelated, so that the most correlated subspaces should retain class information and exclude the noise. Following \citet{Wang_15b}, we extract a low-dimensional projection of the view~1 images with each algorithm, run spectral clustering to partition the splits into $10$ classes (with clustering parameters tuned as in \citep{Wang_15b}), and compare the clustering with ground-truth labels and report the clustering accuracy. We also train a one-vs.-one linear SVM \citep{ChangLin11a} on the projections with highest cluster accuracy for each algorithm (we reveal labels of 10\% of the training set for fast SVM training) and report the classification error rates. The tuning procedure is as for XRMB except that we now select the projection dimensionality from $\{10,20,30\}$. For NCCA/PLCCA we first reduce dimensionality to $100$ by PCA for density estimation and exact nearest neighbor search, and use a randomized algorithm \citep{Halko_11b} to compute the SVD of the $450K\times 450K$ matrix $\SS$; for RKCCA/NKCCA we use an approximation rank of $5000$; for DCCA we use $3$ ReLU hidden layers of $1500$ units in each view and train with stochastic optimization of minibatch size $4500$. Clustering and classification results on the original 784D view~1 inputs are recorded as the baseline. Table~\ref{t:mnist} shows the clustering accuracy and classification error rates on the test set, as well as training run times, and Figure~\ref{f:mnist_emb} shows t-SNE embeddings \citep{MaatenHinton08a} of several algorithms with their optimal hyper-parameters. NCCA and DCCA achieve near perfect class separation.% by implicitly inferring class information using two views of noisy inputs.

\paragraph{Discussion}
Several points are worth noting regarding the experiments.  First, the computation for NCCA and PLCCA is dominated by the exact kNN search; approximate search~\citep{arya1998optimal,andoni2006near} should make NCCA/PLCCA much more efficient.  Second, we have not explored the space of choices for density estimates; alternative choices, such as adaptive KDE~\citep{terrell1992}, could also further improve performance.  Our current choice of KDE would seem to require large training sets for high-dimensional problems.  Indeed, with less training data we do observe a drop in performance, but NCCA still outperforms KCCA; for example, using a 50K subset of the MNIST training set---an order of magnitude less data---the classification error rates when using FKCCA/NKCCA/DCCA/NCCA are 5.9/5.2/2.9/4.7\%.

\vspace{-.3in}
\section{Conclusion}
\vspace{-.07in}
We have presented closed-form solutions to the nonparametric CCA (NCCA) and 
% (KL)
%partially-linear 
partially linear CCA (PLCCA) problems. As opposed to kernel CCA, which restricts the nonparametric projections to lie in a predefined RKHS, we have addressed the unconstrained setting. We have shown that the optimal nonparametric projections can be obtained from the SVD of a kernel defined via the pointwise mutual information between the views. 
% (KL)
This leads to a simple algorithm that outperforms KCCA and matches deep CCA on multiple datasets, while being more computationally efficient than either for moderate-sized data sets.  Future work includes leveraging approximate nearest neighbor search and alternative density estimates. 
%We specifically demonstrated these advantages on several challenging real-world datasets, as well as on a synthetic experiment. 
%\klcomment{added last sentence to address Weiran's suggestion}

\appendix
%\section{Proofs}

\section{Proof of Lemma~\ref{lem:gy_given_fx}}
\label{sec:appendixConsEst} %\wwcomment{$f\rightarrow \f$, and $g\rightarrow \g$?}

Let the eigen-decomposition of the second-order moment of $\expxi{\f(X)|Y}$ be $\expyi{ \expxi{\f(X)|Y}\expxi{\f(X)|Y}^\top }=\A\D\A^\top$ and define $U=\A^\top \expxi{\f(X)|Y}$ and $\tilde{\g}(Y)=\A^\top \g(Y)$. Then the objective in~\eqref{e:ncca} can be written as $\expxyi{\f(X)^\top \g(Y)}=\expyi{\expxi{\f(X)|Y}^\top \g(Y)}=\expyi{(\A^\top\expxi{\f(X)|Y})^\top (\A^\top \g(Y))}=\expyi{U^\top\tilde{\g}(Y)}$. Similarly, the constraint $\I=\expyi{\g(Y)\g(Y)^\top}$ can be expressed as $\I=\A^\top\A=\expyi{(\A^\top \g(Y)) (\A^\top \g(Y))^\top}=\expyi{\tilde{\g}(Y) \tilde{\g}(Y)^\top}$. Therefore, the optimization problem \eqref{e:ncca} can be written in terms of $\tilde{\g}$ as
\begin{align}
\max_{\tilde{\g}}\;  \expy{U^\top \tilde{\g}(Y)} \quad \text{s.t.} \quad  \expy{\tilde{\g}(Y)\tilde{\g}(Y)^\top}=\I. %=\A^\top \expy{{\g}_y{\g}_y^\top} \A=\I.  \nonumber
\end{align}
Our objective is the sum of correlations in all $L$ dimensions. Let us consider the correlation in the $j$th dimension. From the Cauchy-Schwartz inequality, we have
\begin{gather*}
\expy{U_j \tilde{g}_j(Y)}\le \sqrt{ \expy{U_j^2} \expy{ \tilde{g}_j(Y)^2} }=\sqrt{ \expy{U_j^2}} %\sqrt{D_{jj}} \\
\end{gather*}
with equality if and only if $\tilde{g}_{j}(Y)=c_j U_j$ for some scalar $c_j$ with probability 1. Note that choosing each $\tilde{g}_j(Y)$ to be proportional to $U_j$ is valid, since the dimensions of $U$ are uncorrelated (as $\expyi{UU^\top}=\A^\top \expy{\expxi{\f(X)|Y}\expxi{\f(X)|Y}^T} \A=\D$). In order for each $\tilde{g}_j(Y)$ to have unit second order moment, we must have $c_j=1/\sqrt{\expyi{U_j^2}}=1/\sqrt{\D_{jj}}$. Therefore, $\tilde{\g}(Y)=\D^{-1/2} U$ so that ${\g}(Y) = \A \D^{-\frac{1}{2}} \A^\top U = (\expyi{ \expxi{\f(X)|Y}\expxi{\f(X)|Y}^\top })^{-1/2} \expxi{\f(X)|Y}$, proving the lemma.

%\subsection{Maximizing the square-root of a quadratic form}
%\label{sec:appendixMaxSqrt}
%%\begin{lemma}\label{lem:eig}
%%Let $\K$ be any $D\times D$ symmetric positive semi-definite matrix. We are interested in the optimum of the problem
%%  \begin{align}\label{e:eig}
%%    \max_{\F\in \bbR^{D\times L}} \; \trace{\left( \F^\top \K \F \right)^{\frac{1}{2}}} \qquad \text{s.t.} \quad \F^\top \F=\I
%%  \end{align}
%%  is achieved by the top $L$ eigenvectors of $\K$.
%%\end{lemma}
%%\begin{proof}
%Since the gradient of $\tracei{\C^{1/2}}$ is $\tfrac{1}{2} \tracei{\C^{-1/2}}$ (see, e.g. \cite[Theorem~1]{Lewis96a}), we have that $\partial \tracei{( \tilde{\W}^\top \K \tilde{\W} )^{1/2}}/\partial \tilde{\W} = \K\tilde{\W}(\tilde{\W}^\top \K \tilde{\W})^{-1/2}$. Using this derivative within the KKT theorem, we can show that the maximum of $\{\tracei{( \tilde{\W}^\top \K \tilde{\W} )^{1/2}}:\tilde{\W}^\top\tilde{\W}=\I\}$ is attained by a $\tilde{\W}$ which satisfies $\K\tilde{\W}(\tilde{\W}^\top \K \tilde{\W})^{-1/2}=\tilde{\W}\Lambda$, where $\Lambda$ is a matrix containing the Lagrange multipliers associated with the constraints. Consequently, $\K\tilde{\W}=\tilde{\W}\Lambda (\tilde{\W}^\top \K \tilde{\W})^{1/2}$ and this equality is satisfied if $\tilde{\W}$ contains eigenvectors of $\K$ (in which case $(\tilde{\W}^\top \K \tilde{\W})^{1/2}$ is diagonal). It is then easy to see that the optimal $\tilde{\W}$ contains the eigenvectors associated with the largest $L$ eigenvalues of $\K$.
%%\end{proof} 

\subsubsection*{Acknowledgement}
Thanks to Nathan Srebro, Ryota Tomioka, and Yochai Blau for fruitful discussions.  This research was supported by NSF grant IIS-1321015.  The opinions expressed in this work are those of the authors and do not necessarily reflect the views of the funding agency.

\bibliographystyle{abbrvnat}
\bibliography{icml16b}

\begin{thebibliography}{47}
\providecommand{\natexlab}[1]{#1}
\providecommand{\url}[1]{\texttt{#1}}
\expandafter\ifx\csname urlstyle\endcsname\relax
  \providecommand{\doi}[1]{doi: #1}\else
  \providecommand{\doi}{doi: \begingroup \urlstyle{rm}\Url}\fi

\bibitem[Akaho(2001)]{Akaho01a}
S.~Akaho.
\newblock A kernel method for canonical correlation analysis.
\newblock In \emph{Proceedings of the International Meeting of the Psychometric
  Society (IMPS2001)}. Springer-Verlag, 2001.

\bibitem[Andoni and Indyk(2006)]{andoni2006near}
A.~Andoni and P.~Indyk.
\newblock Near-optimal hashing algorithms for approximate nearest neighbor in
  high dimensions.
\newblock In \emph{Foundations of Computer Science, 2006. FOCS'06. 47th Annual
  IEEE Symposium on}, pages 459--468. IEEE, 2006.

\bibitem[Andrew et~al.(2013)Andrew, Arora, Bilmes, and Livescu]{Andrew_13a}
G.~Andrew, R.~Arora, J.~Bilmes, and K.~Livescu.
\newblock Deep canonical correlation analysis.
\newblock In \emph{Proc. of the 30th Int. Conf. Machine Learning (ICML 2013)},
  pages 1247--1255, 2013.

\bibitem[Arora and Livescu(2012)]{AroraLivesc12a}
R.~Arora and K.~Livescu.
\newblock Kernel {CCA} for multi-view learning of acoustic features using
  articulatory measurements.
\newblock In \emph{Symposium on Machine Learning in Speech and Language
  Processing (MLSLP)}, 2012.

\bibitem[Arora and Livescu(2013)]{AroraLivesc13a}
R.~Arora and K.~Livescu.
\newblock Multi-view {CCA}-based acoustic features for phonetic recognition
  across speakers and domains.
\newblock In \emph{Proc. of the IEEE Int. Conf. Acoustics, Speech and Sig.
  Proc. (ICASSP'13)}, 2013.

\bibitem[Arya et~al.(1998)Arya, Mount, Netanyahu, Silverman, and
  Wu]{arya1998optimal}
S.~Arya, D.~M. Mount, N.~S. Netanyahu, R.~Silverman, and A.~Y. Wu.
\newblock An optimal algorithm for approximate nearest neighbor searching fixed
  dimensions.
\newblock \emph{Journal of the ACM (JACM)}, 45\penalty0 (6):\penalty0 891--923,
  1998.

\bibitem[Bach and Jordan(2002)]{BachJordan02a}
F.~R. Bach and M.~I. Jordan.
\newblock Kernel independent component analysis.
\newblock \emph{Journal of Machine Learning Research}, 3:\penalty0 1--48, 2002.

\bibitem[Bach and Jordan(2005)]{BachJordan05a}
F.~R. Bach and M.~I. Jordan.
\newblock A probabilistic interpretation of canonical correlation analysis.
\newblock Technical Report 688, Dept. of Statistics, University of California,
  Berkeley, 2005.

\bibitem[Balakrishnan et~al.(2012)Balakrishnan, Puniyani, and
  Lafferty]{Balakr_12a}
S.~Balakrishnan, K.~Puniyani, and J.~Lafferty.
\newblock Sparse additive functional and kernel {CCA}.
\newblock In \emph{Proc. of the 29th Int. Conf. Machine Learning (ICML 2012)},
  pages 911--918, 2012.

\bibitem[Bolla(2013)]{bolla2013}
M.~Bolla.
\newblock \emph{Spectral Clustering and Biclustering: Learning Large Graphs and
  Contingency Tables}.
\newblock John Wiley \& Sons, 2013.

\bibitem[Boots and Gordon(2012)]{BootsGordon12a}
B.~Boots and G.~Gordon.
\newblock Two manifold problems with applications to nonlinear system
  identification.
\newblock In \emph{Proc. of the 29th Int. Conf. Machine Learning (ICML 2012)},
  pages 623--630, 2012.

\bibitem[Borga(2001)]{Borga01a}
M.~Borga.
\newblock Canonical correlation: A tutorial.
\newblock 2001.

\bibitem[Breiman and Friedman(1985)]{Breiman85}
L.~Breiman and J.~H. Friedman.
\newblock Estimating optimal transformations for multiple regression and
  correlation.
\newblock \emph{Journal of the American statistical Association}, 80\penalty0
  (391):\penalty0 580--598, 1985.

\bibitem[Chang and Lin(2011)]{ChangLin11a}
C.-C. Chang and C.-J. Lin.
\newblock {LIBSVM}: A library for support vector machines.
\newblock \emph{ACM Trans. Intelligent Systems and Technology}, 2\penalty0
  (3):\penalty0 27, 2011.

\bibitem[Chechik et~al.(2005)Chechik, Globerson, Tishby, and
  Weiss]{Chechik_05a}
G.~Chechik, A.~Globerson, N.~Tishby, and Y.~Weiss.
\newblock Information bottleneck for {Gaussian} variables.
\newblock \emph{Journal of Machine Learning Research}, 6:\penalty0 165--188,
  2005.

\bibitem[de~Sa(2005)]{Sa05a}
V.~de~Sa.
\newblock Spectral clustering with two views.
\newblock In \emph{Workshop on Learning with Multiple Views (ICML'05)}, pages
  20--27, 2005.

\bibitem[Dhillon et~al.(2011)Dhillon, Foster, and Ungar]{Dhillon_11b}
P.~Dhillon, D.~Foster, and L.~Ungar.
\newblock Multi-view learning of word embeddings via {CCA}.
\newblock In \emph{Advances in Neural Information Processing Systems (NIPS)},
  volume~24, pages 199--207, 2011.

\bibitem[Eagleson(1964)]{eagleson1964polynomial}
G.~Eagleson.
\newblock Polynomial expansions of bivariate distributions.
\newblock \emph{The Annals of Mathematical Statistics}, pages 1208--1215, 1964.

\bibitem[Eldar and Oppenheim(2003)]{Eldar_03}
Y.~C. Eldar and A.~V. Oppenheim.
\newblock {MMSE} whitening and subspace whitening.
\newblock \emph{IEEE Trans. Info. Theory}, 49\penalty0 (7):\penalty0
  1846--1851, 2003.

\bibitem[Georgescu et~al.(2003)Georgescu, Shimshoni, and Meer]{Georges_03a}
B.~Georgescu, I.~Shimshoni, and P.~Meer.
\newblock Mean shift based clustering in high dimensions: {A} texture
  classification example.
\newblock In \emph{Proc. 9th Int. Conf. Computer Vision (ICCV'03)}, pages
  456--463, Nice, France, Oct.~14--17 2003.

\bibitem[Gong et~al.(2014)Gong, Wang, Hodosh, Hockenmaier, and
  Lazebnik]{gong2014improving}
Y.~Gong, L.~Wang, M.~Hodosh, J.~Hockenmaier, and S.~Lazebnik.
\newblock Improving image-sentence embeddings using large weakly annotated
  photo collections.
\newblock In \emph{European Conference on Computer Vision}, 2014.

\bibitem[Halko et~al.(2011)Halko, Martinsson, Shkolnisky, and
  Tygert]{Halko_11b}
N.~Halko, P.-G. Martinsson, Y.~Shkolnisky, and M.~Tygert.
\newblock An algorithm for the principal component analysis of large data sets.
\newblock \emph{SIAM J. Sci. Comput.}, 33\penalty0 (5):\penalty0 2580--2594,
  2011.

\bibitem[Hannan(1961)]{hannan1961}
E.~J. Hannan.
\newblock The general theory of canonical correlation and its relation to
  functional analysis.
\newblock \emph{Journal of the Australian Mathematical Society}, 2\penalty0
  (02):\penalty0 229--242, 1961.

\bibitem[Hardoon et~al.(2004)Hardoon, Szedmak, and Shawe-Taylor]{Hardoon_04a}
D.~R. Hardoon, S.~Szedmak, and J.~Shawe-Taylor.
\newblock Canonical correlation analysis: An overview with application to
  learning methods.
\newblock \emph{Neural Computation}, 16\penalty0 (12):\penalty0 2639--2664,
  2004.

\bibitem[Hotelling(1936)]{Hotell36a}
H.~Hotelling.
\newblock Relations between two sets of variates.
\newblock \emph{Biometrika}, 28\penalty0 (3/4):\penalty0 321--377, 1936.

\bibitem[Kumar et~al.(2011)Kumar, Rai, and {Daum{\'e}~III}]{Kumar_11a}
A.~Kumar, P.~Rai, and H.~{Daum{\'e}~III}.
\newblock Co-regularized multi-view spectral clustering.
\newblock In \emph{Advances in Neural Information Processing Systems (NIPS)},
  volume~24, pages 1413--1421, 2011.

\bibitem[Lai and Fyfe(2000)]{LaiFyfe00a}
P.~L. Lai and C.~Fyfe.
\newblock Kernel and nonlinear canonical correlation analysis.
\newblock \emph{Int. J. Neural Syst.}, 10\penalty0 (5):\penalty0 365--377,
  2000.

\bibitem[Lancaster(1958)]{lancaster1958}
H.~Lancaster.
\newblock The structure of bivariate distributions.
\newblock \emph{The Annals of Mathematical Statistics}, pages 719--736, 1958.

\bibitem[{LeCun} et~al.(1998){LeCun}, Bottou, Bengio, and Haffner]{Lecun_98a}
Y.~{LeCun}, L.~Bottou, Y.~Bengio, and P.~Haffner.
\newblock Gradient-based learning applied to document recognition.
\newblock \emph{Proc. IEEE}, 86\penalty0 (11):\penalty0 2278--2324, 1998.

\bibitem[Lederman and Talmon(2014)]{LedermTalmon14a}
R.~R. Lederman and R.~Talmon.
\newblock Common manifold learning using alternating-diffusion.
\newblock Technical Report YALEU/DCS/TR-1497, 2014.

\bibitem[Lopez-Paz et~al.(2014)Lopez-Paz, Sra, Smola, Ghahramani, and
  Schoelkopf]{Lopez_14b}
D.~Lopez-Paz, S.~Sra, A.~Smola, Z.~Ghahramani, and B.~Schoelkopf.
\newblock Randomized nonlinear component analysis.
\newblock In \emph{Proc. of the 31st Int. Conf. Machine Learning (ICML 2014)},
  pages 1359--1367, 2014.

\bibitem[Makur et~al.(2015)Makur, Kozynski, Huang, and Zheng]{Makur_15a}
A.~Makur, F.~Kozynski, S.-L. Huang, and L.~Zheng.
\newblock An efficient algorithm for information decomposition and extraction.
\newblock In \emph{53rd Annual Allerton Conference on Communication, Control,
  and Computing}, 2015.

\bibitem[Mardia et~al.(1979)Mardia, Kent, and Bibby]{mardia_79}
K.~V. Mardia, J.~T. Kent, and J.~M. Bibby.
\newblock \emph{Multivariate Analysis}.
\newblock Academic Press, 1979.

\bibitem[Melzer et~al.(2001)Melzer, Reiter, and Bischof]{Melzer_01a}
T.~Melzer, M.~Reiter, and H.~Bischof.
\newblock Nonlinear feature extraction using generalized canonical correlation
  analysis.
\newblock In \emph{Proc. of the 11th Int. Conf. Artificial Neural Networks
  (ICANN'01)}, pages 353--360, 2001.

\bibitem[Nadaraya(1964)]{nadaraya1964}
E.~A. Nadaraya.
\newblock On estimating regression.
\newblock \emph{Theory of Probability \& Its Applications}, 9\penalty0
  (1):\penalty0 141--142, 1964.

\bibitem[Nair and Hinton(2010)]{NairHinton10a}
V.~Nair and G.~E. Hinton.
\newblock Rectified linear units improve restricted {Boltzmann} machines.
\newblock In \emph{Proc. of the 27th Int. Conf. Machine Learning (ICML 2010)},
  pages 807--814, June~21--25 2010.

\bibitem[Ozakin and Gray(2009)]{OzakinGray09a}
A.~Ozakin and A.~Gray.
\newblock Submanifold density estimation.
\newblock In \emph{Advances in Neural Information Processing Systems (NIPS)},
  volume~22, pages 1375--1382, 2009.

\bibitem[Terrell and Scott(1992)]{terrell1992}
G.~R. Terrell and D.~W. Scott.
\newblock Variable kernel density estimation.
\newblock \emph{The Annals of Statistics}, pages 1236--1265, 1992.

\bibitem[van~der Maaten and Hinton(2008)]{MaatenHinton08a}
L.~J.~P. van~der Maaten and G.~E. Hinton.
\newblock Visualizing data using $t$-{SNE}.
\newblock \emph{Journal of Machine Learning Research}, 9:\penalty0 2579--2605,
  2008.

\bibitem[Wang et~al.(2012)Wang, Jiang, Wang, Zhou, and Tu]{Wang_12c}
B.~Wang, J.~Jiang, W.~Wang, Z.-H. Zhou, and Z.~Tu.
\newblock Unsupervised metric fusion by cross diffusion.
\newblock In \emph{Proc. of the 2012 IEEE Computer Society Conf. Computer
  Vision and Pattern Recognition (CVPR'12)}, pages 2997--3004, 2012.

\bibitem[Wang et~al.(2015{\natexlab{a}})Wang, Arora, Livescu, and
  Bilmes]{Wang_15a}
W.~Wang, R.~Arora, K.~Livescu, and J.~Bilmes.
\newblock Unsupervised learning of acoustic features via deep canonical
  correlation analysis.
\newblock In \emph{Proc. of the IEEE Int. Conf. Acoustics, Speech and Sig.
  Proc. (ICASSP'15)}, 2015{\natexlab{a}}.

\bibitem[Wang et~al.(2015{\natexlab{b}})Wang, Arora, Livescu, and
  Bilmes]{Wang_15b}
W.~Wang, R.~Arora, K.~Livescu, and J.~Bilmes.
\newblock On deep multi-view representation learning.
\newblock In \emph{Proc. of the 32st Int. Conf. Machine Learning (ICML 2015)},
  2015{\natexlab{b}}.

\bibitem[Watson(1964)]{watson1964}
G.~S. Watson.
\newblock Smooth regression analysis.
\newblock \emph{Sankhy{\=a}: The Indian Journal of Statistics, Series A}, pages
  359--372, 1964.

\bibitem[Westbury(1994)]{Westbur94a}
J.~R. Westbury.
\newblock \emph{X-Ray Microbeam Speech Production Database User's Handbook
  Version 1.0}, 1994.

\bibitem[Williams and Seeger(2001)]{WilliamSeeger01a}
C.~K.~I. Williams and M.~Seeger.
\newblock Using the {Nystr{\"o}m} method to speed up kernel machines.
\newblock In \emph{Advances in Neural Information Processing Systems (NIPS)},
  volume~13, pages 682--688, 2001.

\bibitem[Witten and Tibshirani(2009)]{Witten09}
D.~M. Witten and R.~J. Tibshirani.
\newblock Extensions of sparse canonical correlation analysis with applications
  to genomic data.
\newblock \emph{Statistical applications in genetics and molecular biology},
  8\penalty0 (1):\penalty0 1--27, 2009.

\bibitem[Zhou and Burges(2007)]{ZhouBurges07a}
D.~Zhou and C.~J.~C. Burges.
\newblock Spectral clustering and transductive learning with multiple views.
\newblock In \emph{Proc. of the 24th Int. Conf. Machine Learning (ICML'07)},
  pages 1159--1166, 2007.

\end{thebibliography}

\end{document}

\twocolumn[

\icmltitle{Supplementary materials for \\Nonparametric Canonical Correlation Analysis}

\icmlauthor{Tomer Michaeli}{tomer.m@technion.ac.il}
\icmladdress{Technion--Israel Institute of Technology,
            Haifa, Israel}
\icmlauthor{Weiran Wang}{weiranwang@ttic.edu}
\icmladdress{TTI-Chicago, 
  Chicago, IL 60637, USA}
\icmlauthor{Karen Livescu}{klivescu@ttic.edu}
\icmladdress{TTI-Chicago, 
  Chicago, IL 60637, USA}


\vskip 0.3in
]

\section{Proof of Lemma~3.1}
\label{sec:appendixConsEst}

\begin{proof}
Let the eigen-decomposition of the second-order moment of $\expxi{\f(X)|Y}$ be $\expyi{ \expxi{\f(X)|Y}\expxi{\f(X)|Y}^\top }=\A\D\A^\top$ and define $U=\A^\top \expxi{\f(X)|Y}$ and $\tilde{\g}(Y)=\A^\top \g(Y)$. Then the objective in~(3) % \eqref{e:ncca} 
can be written as $\expxyi{\f(X)^\top \g(Y)}=\expyi{\expxi{\f(X)|Y}^\top \g(Y)}=\expyi{(\A^\top\expxi{\f(X)|Y})^\top (\A^\top \g(Y))}=\expyi{U^\top\tilde{\g}(Y)}$. Similarly, the constraint $\I=\expyi{\g(Y)\g(Y)^\top}$ can be expressed as $\I=\A^\top\A=\expyi{(\A^\top \g(Y)) (\A^\top \g(Y))^\top}=\expyi{\tilde{\g}(Y) \tilde{\g}(Y)^\top}$. Therefore, the optimization problem~(3) % \eqref{e:ncca} 
can be written in terms of $\tilde{\g}$ as
\begin{align*}
\max_{\tilde{\g}}\;  \expy{U^\top \tilde{\g}(Y)} \quad \text{s.t.} \quad  \expy{\tilde{\g}(Y)\tilde{\g}(Y)^\top}=\I. %=\A^\top \expy{{\g}_y{\g}_y^\top} \A=\I.  \nonumber
\end{align*}
Our objective is the sum of correlations in all $L$ dimensions. Let us consider the correlation in the $j$th dimension. From the Cauchy-Schwartz inequality, we have
\begin{gather*}
\expy{U_j \tilde{g}_j(Y)}\le \sqrt{ \expy{U_j^2} \expy{ \tilde{g}_j(Y)^2} }=\sqrt{ \expy{U_j^2}} %\sqrt{D_{jj}} \\
\end{gather*}
with equality if and only if $\tilde{g}_{j}(Y)=c_j U_j$ for some scalar $c_j$ with probability 1. Note that choosing each $\tilde{g}_j(Y)$ to be proportional to $U_j$ is valid, since the dimensions of $U$ are uncorrelated (as $\expyi{UU^\top}=\A^\top \expy{\expxi{\f(X)|Y}\expxi{\f(X)|Y}^T} \A=\D$). In order for each $\tilde{g}_j(Y)$ to have unit second order moment, we must have $c_j=1/\sqrt{\expyi{U_j^2}}=1/\sqrt{\D_{jj}}$. Therefore, $\tilde{\g}(Y)=\D^{-1/2} U$ so that ${\g}(Y) = \A \D^{-\frac{1}{2}} \A^\top U = (\expyi{ \expxi{\f(X)|Y}\expxi{\f(X)|Y}^\top })^{-1/2} \expxi{\f(X)|Y}$, proving the lemma.
\end{proof}

% \bibliographystyle{icml2016}
% \bibliography{icml16b}